%% file: main.tex
\PassOptionsToPackage{table}{xcolor}
\documentclass{article}
\usepackage[preprint]{acl}
\usepackage{times}
\input{math_commands.tex}
\usepackage{hyperref}
\usepackage{url}
\usepackage{booktabs}
\usepackage{xcolor}
\usepackage{amsfonts}
\usepackage{amsmath}
\usepackage{microtype}
\usepackage{graphicx}
\usepackage{float}
\usepackage{needspace}
\usepackage{placeins}
\usepackage[T1]{fontenc}
\usepackage[utf8]{inputenc}

\newcommand{\NA}{--}

\title{BatchParScale: Sparse Batch-Mixing for Sequence Models}
\author{%
  Hans Peter Lyngsøe Raaschou-jensen\textsuperscript{1}
  \quad M. Onurcan Kaya\textsuperscript{2}
  \quad Desmond Elliott\textsuperscript{1} \\
  \textsuperscript{1}University of Copenhagen \\
  \textsuperscript{2}Technical University of Denmark
}
\hypersetup{%
  pdftitle={BatchParScale: Sparse Batch-Mixing for Sequence Models},
  pdfauthor={Hans Peter Lyngsøe Raaschou-jensen, M. Onurcan Kaya, Desmond Elliott}
}

\begin{document}
\maketitle

\begin{abstract}
\noindent Recent test-time scaling methods improve model performance by increasing inference-time compute.
One practical form of this is parallel scaling, where a model samples multiple independent reasoning streams and aggregates their answers using methods like voting, reranking, or verification.
While effective, this approach leaves the streams uncoordinated, and a discovery made in one rollout cannot influence any other. Multi-Sequence Verifiers jointly score candidate answers after generation, but the reasoning streams remain independent during inference.
ParScale takes a step toward model-internal parallel scaling by learning to aggregate multiple forward passes, but its streams still do not exchange information while processing a sequence.
We introduce \method{}, an attention mechanism that allows parallel streams to exchange information through a small set of communication positions.
We evaluate \method{} on a new Parallel Needle-in-a-Haystack retrieval benchmark and in pretraining, fine-tuning, and cooperative multi-agent reinforcement learning.
Across these settings, \method{} outperforms \parscale{} on language-model loss and remains competitive on coordination-sensitive MARL tasks.
Masked pretraining runs beat an unmasked single-sequence baseline despite local sequence corruption, and LoRA fine-tuning shows the mechanism can be added to a pretrained Transformer.
At the same time, fact retrieval performance for pretrained and finetuned \method{} models remains unclear.
\end{abstract}

\begin{figure*}[t]
\centering
\includegraphics[width=\textwidth]{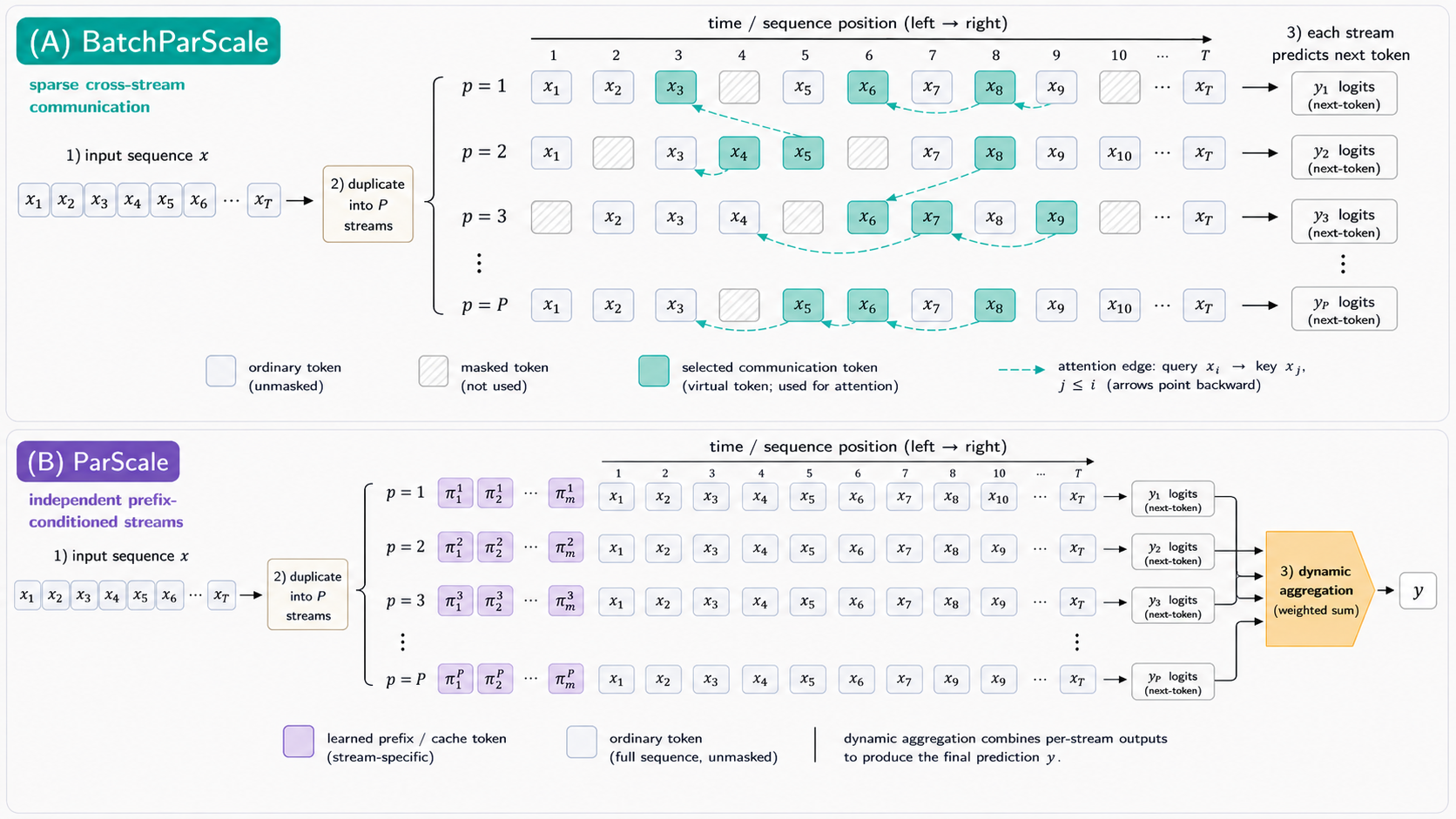}
\caption{Overview of \method{}. (A) \method{} lets parallel streams share selected hidden states through sparse communication positions. (B) \parscale{} streams interact only at final aggregation.}
\label{fig:main-overview}
\end{figure*}

\section{Introduction}

\noindent Parallel test-time scaling has become a practical way to spend more computation at inference time.
The common recipe is to sample several candidate reasoning traces, then choose among them with voting, reranking, or verification \citep{wang2023selfconsistency,snell2024testtime,cobbe2021trainingverifiers,lightman2023letsverifystep}.
This can improve answer coverage, but the parallel traces in one rollout cannot reuse information from another.
For example, if different traces explore different proof strategies, an intermediate lemma found in one trace is unavailable to the others until final aggregation.
Parallel scaling therefore increases the number of computations, but it does not make those computations collaborative.

\noindent Final aggregation and verification can compare completed traces, but they do not let one stream change another stream's computation during inference.
This work asks whether sequence models can occupy a tunable middle ground between parallel and sequential scaling: keeping the diversity of parallel streams while allowing intentional communication during the forward pass.
We introduce batch-mixing as a class of sequence models in which streams can attend to selected hidden states from their peers.
Our concrete instantiation, \method{}, restricts this exchange to a small set of communication positions, so most tokens remain ordinary per-stream causal computation.

\noindent We evaluate this idea from several angles.
Parallel Needle-in-a-Haystack tests whether a stream without local evidence can recover information from another stream.
TinyStories pretraining and FineWeb-Edu LoRA continued pretraining test whether the mechanism improves ordinary language-model objectives and masked-token recovery.
Parallel QA-style probes test whether those gains correspond to reliable transfer of specific facts across streams.
Cooperative multi-agent reinforcement learning tests whether the same communication primitive can support learned coordination.
Finally, because sparse attention patterns can be asymptotically attractive but slow in practice, we analyze the direct implementation and an efficient two-pass version of \method{}.

\noindent \method{} can learn controlled cross-stream retrieval, improves aggregate language-model loss in our small-scale comparisons, and improves masked-token recovery when neighboring streams contain redundant context.
LoRA continued pretraining shows that the mechanism can be added to a pretrained Transformer, and MARL results show competitive performance on several coordination-sensitive tasks.
However, aggregate loss improvements and masked-token recovery do not imply reliable transfer of specific facts to streams without local evidence.
This failure mode is clearest in PNIAH and PMHQA, where for both pretrained and finetuned models performance remains close to random despite the model being able to solve the task perfectly when information remains in a single stream. On the other hand, performance on parallel-WikiHop remains meaningfully above random. This motivates treating better data augmentation and training objectives as open problems for future research.

\noindent Our contributions are:
\begin{itemize}
\item We formulate \xattn as a sequence-modeling primitive for communication between parallel streams, and instantiate it as \method{}.
\item We introduce Parallel Needle-in-a-Haystack (PNIAH) and parallel QA/MHQA (PQA/PMHQA)-style benchmarks for measuring whether information reaches streams that do not contain the evidence locally.
\item We evaluate \method{} across language-model pretraining, LoRA continued pretraining, PNIAH and PMHQA, and cooperative MARL, finding that our architecture outperforms \parscale{} on sequence modelling, but performs poorly on downstream PNIAH and PMHQA-style benchmarks.
\item We produce an efficient two-pass implementation that makes \method{} communication hardware-friendly.
\end{itemize}

\section{Related Work}

\subsection{Parallel Reasoning and Test-Time Scaling}

\noindent Recent work treats inference-time computation as a scaling axis for language models. These scaling methods allocate additional computation to sampling, verification, or refinement, and have been shown in certain cases to improve performance more efficiently than increasing model size \citep{snell2024testtime}.
The most common parallel form is to sample multiple independent candidate solutions and aggregate them.
Self-consistency uses this idea for chain-of-thought reasoning by sampling multiple reasoning paths and selecting the most consistent answer \citep{wang2023selfconsistency}.
Large Language Monkeys studies this repeated-sampling regime at much larger scale, showing that solution coverage can improve over many samples, especially in domains with reliable automatic verification \citep{brown2024largelanguagemonkeys}.
However, these approaches leave the underlying computations independent: each rollout is generated without access to the intermediate states or discoveries of the others.

\noindent Multi-Sequence Verifiers address this limitation by jointly processing a set of candidate answers instead of scoring each candidate independently \citep{kim2026multisequenceverifiers}.
They improve best-of-\(N\) selection and early stopping through an auxiliary verifier, but the reasoning streams themselves still do not communicate while generating.

\noindent ParScale scales inference differently, by running parallel prefix-conditioned forward passes over the same sequence rather than sampling independent outputs \citep{chen2025parscale}.
It duplicates the input across \(P\) streams that share a Transformer backbone; each stream prepends fixed, input-independent learned prefix tokens that diversify hidden states across streams.
Their outputs are combined only at a final dynamic aggregation step, and the streams do not interact during the forward pass.

\subsection{Sparse and Latent Attention}

\noindent The Transformer introduced self-attention as the core sequence-modeling primitive \citep{vaswani2017attention}.
Encoder-decoder attention, in which queries come from one sequence and keys and values come from another, provides an early form of cross-sequence information exchange \citep{bahdanau2015neural,vaswani2017attention}.
Subsequent work has studied sparse and latent variants of these attention structures.
Sparse Transformer, Longformer, and BigBird reduce the overhead of dense self-attention through local, strided, random, or global connectivity patterns \citep{child2019sparse,beltagy2020longformer,zaheer2020bigbird}.
Longformer furthermore designates selected positions as global-attention sites and gives them dedicated query, key, and value projections rather than reusing the ordinary token projections.
Set Transformer and Perceiver-style architectures introduce learned latent slots or inducing points that mediate attention over large collections of inputs \citep{lee2019settransformer,jaegle2021perceiver,jaegle2021perceiverio}.
Our communication positions play a similar inducing-point role, but they are embedded inside otherwise independent streams and are constrained by causal sequence order.

\subsection{Multi-Agent Reinforcement Learning}

\noindent Multi-agent reinforcement learning studies settings in which multiple agents learn policies while interacting through a shared environment \citep{tan1993multiagent,littman1994markov,busoniu2008survey}.
A common paradigm is centralized training with decentralized execution, where agents may use global information during training but act from local observations at test time \citep{oliehoek2008decpomdp,lowe2017maddpg}.
MADDPG introduced an actor-critic approach for mixed cooperative-competitive environments and popularized the Multi-Agent Particle Environment (MPE) benchmarks \citep{lowe2017maddpg}.
MAPPO later showed that a carefully tuned PPO implementation is a strong and simple baseline for cooperative MARL tasks \citep{schulman2017ppo,yu2022mappo}.

\noindent A related line of work learns differentiable communication channels between separate actors.
CommNet lets agents condition on an average of other agents' hidden states \citep{sukhbaatar2016commnet}; IC3Net adds learned gates that control when agents communicate \citep{singh2019ic3net}; and TarMAC uses attention to make communication targeted across senders and receivers \citep{das2019tarmac}.
These methods establish continuous hidden-state exchange as a trainable component of cooperative decision-making systems.
Our MARL experiments place \xattn{} in this lineage by treating each agent's history as a stream and restricting cross-agent reads to selected communication positions.

\noindent Recent work has also applied sequence models to MARL.
Multi-Agent Transformer formulates cooperative MARL as sequence modeling over agents and actions \citep{wen2022mat}.
Sable further studies scalable sequence models for MARL, emphasizing performance, memory efficiency, and scalability \citep{mahjoub2025sable}.
JAX-based environment suites have also made large batched RL experiments easier to run: Jumanji provides scalable single-agent environments and baselines \citep{bonnet2024jumanji}, while JaxMARL extends this style of infrastructure to multi-agent environments and algorithms \citep{rutherford2023jaxmarl}.
Brax provides the differentiable physics backend used by continuous-control multi-agent environments such as MABrax \citep{freeman2021brax}.

\section{Method}

\subsection{ParScale}

\noindent The model architecture most similar to our approach is \parscale{} \citep{chen2025parscale}. Given an input sequence
\[
x = (x_1,\ldots,x_n),
\]
\parscale{} duplicates the sequence into \(P \in \mathbb{N}\) parallel streams.
Each stream is processed by the same Transformer backbone \(f_\theta\), but the streams are distinguished by learned prefix key-value caches.
For stream \(p \in \{1,\ldots,P\}\) and layer \(\ell\), let
\[
\Pi^{(\ell)}_p =
\left(
K^{(\ell)}_{p,\mathrm{pre}},
V^{(\ell)}_{p,\mathrm{pre}}
\right)
\]
denote the stream-specific learned prefix cache, with prefix length \(q \in \mathbb{N}\).
For one attention head we have:
\[
K^{(\ell)}_{p,\mathrm{pre}},
V^{(\ell)}_{p,\mathrm{pre}}
\in
\mathbb{R}^{q \times d_h}.
\]
If the ordinary hidden states in stream \(p\) produce queries, keys, and values
\[
Q^{(\ell)}_p,\quad K^{(\ell)}_{p,x},\quad V^{(\ell)}_{p,x}
\in \mathbb{R}^{n \times d_h}
\]
then the attention operation in \parscale{} is schematically
\[
\begin{aligned}
\widetilde{K}^{(\ell)}_p =
\operatorname{Concat}
\left[
K^{(\ell)}_{p,\mathrm{pre}};
K^{(\ell)}_{p,x}
\right],\\
\widetilde{V}^{(\ell)}_p =
\operatorname{Concat}
\left[
V^{(\ell)}_{p,\mathrm{pre}};
V^{(\ell)}_{p,x}
\right],
\end{aligned}
\]
\[
\operatorname{Attn}
\left(
Q^{(\ell)}_p,\,
\widetilde{K}^{(\ell)}_p,\,
\widetilde{V}^{(\ell)}_p
\right).
\]
Thus, all streams receive the same token sequence \(x\), but stream \(p\) attends with its own learned prefix cache at every layer.
We can summarize the resulting model as
\[
g_\theta(x)
=
\sum_{p=1}^{P}
w_p f_\theta(x_p),
\]
where \(x_p\) denotes the \(p\)-th prefix-conditioned stream and \(w_p\) is its aggregation weight.
The backbone parameters \(\theta\) are shared across all streams; only the prefix caches and aggregation parameters are stream-specific.

\noindent The aggregation weights are computed dynamically from the stream outputs, using an MLP
\[
h : \mathbb{R}^{d_o \times P} \to \mathbb{R}^{P}
\]
that maps stream outputs to \(P\) logits.

\noindent The aggregation weights are computed as
\[
\mathbf{w}
=
\operatorname{Softmax}
\left(
h\left(
\left[
f_\theta(x_1);
\ldots;
f_\theta(x_P)
\right]
\right)
\right).
\]
Training optimizes standard next-token cross-entropy on the aggregated output of \(g_\theta\), so all \(P\) streams contribute to a single prediction at each position rather than receiving separate per-stream losses. To avoid collapse, label smoothing is also applied to the aggregation weights,
\[
w_p \leftarrow w_p(1-\epsilon) + \frac{\epsilon}{P},
\]
which acts as a load-balancing regularizer and prevents a few streams from dominating the mixture early in training.

\noindent \parscale{} increases computation by evaluating \(P\) prefix-conditioned versions of the same sequence through a shared Transformer.
The streams are diverse because they use different learned prefix key-value caches, but they remain independent while processing the sequence.
Their only interaction occurs at the final dynamic aggregation step.
The cost of adding sparse cross-stream communication is analyzed in Section~\ref{sec:perf}.

\subsection{Batch-Mixing}

\noindent Transformer layers usually mix hidden states along two axes. Given
hidden states \(X \in \mathbb{R}^{B \times S \times D}\), the MLP block is a
channel mixer: it applies the same token-wise map over the feature dimension,
\[
X_{b,t,:} \mapsto \operatorname{MLP}(X_{b,t,:}),
\]
without exchanging information across positions or batch elements. Causal
self-attention is a sequence mixer. After projecting hidden states to queries,
keys, and values,
\[
Q,K,V \in \mathbb{R}^{B \times S \times d_h},
\]
the output at position \(t\) in batch element \(b\) is a weighted sum of value
vectors from earlier positions in the same sequence,
\[
\operatorname{Attn}(X)_{b,t,:}
=
\sum_{s \leq t}
\alpha_{b,t,s} V_{b,s,:}.
\]
Thus ordinary causal attention mixes over the sequence index \(s\), but keeps
the batch index fixed.

\noindent Batch-mixing changes the attention mask so that
tokens can also read from other batch elements. In practice, batch-mixing architectures cannot allow all tokens to communicate across the batch, as this quickly becomes computationally intractable.

\subsection{\method{}}

\noindent \parscale{} creates multiple prefix-conditioned streams, but those streams are independent until the final aggregation. 

We now introduce \method{} (BPS), our first batch-mixing architecture in which \(B\) parallel streams are grouped in one forward pass and only designated communication positions may attend across streams.

\noindent A small set of communication positions mediates information flow between otherwise independent streams.
Each stream is processed as its own causal sequence through the shared Transformer, so cross-stream reads are a restricted form of cross-attention where queries come from one sequence and keys and values from another.

\noindent For each original sequence, we create \(P\) streams and group them as a single communication set.
Let \(H^{(\ell)}_p \in \mathbb{R}^{n \times d}\) denote the hidden states of stream \(p\) entering layer \(\ell\).
As in the traditional Transformer, each stream produces
\[
Q^{(\ell)}_p,\quad K^{(\ell)}_{p,x},\quad V^{(\ell)}_{p,x}
\in \mathbb{R}^{n \times d_h}.
\]
Rather than prepend stream-specific prefix caches, we designate a sparse set of \emph{virtual tokens} \(C^{(\ell)}_p \subseteq \{1,\ldots,n\}\) for stream \(p\) at layer \(\ell\).

\noindent Let \(K^{(\ell)}\) and \(V^{(\ell)}\) stack the keys and values of all \(P\) streams,
\[
\begin{aligned}
K^{(\ell)} &=
\left[
K^{(\ell)}_{1,x};\ldots;K^{(\ell)}_{P,x}
\right],
\\
V^{(\ell)} &=
\left[
V^{(\ell)}_{1,x};\ldots;V^{(\ell)}_{P,x}
\right]
\in \mathbb{R}^{Pn \times d_h},
\end{aligned}
\]
so a query in stream \(p\) could in principle attend to any position in any stream. Cross-stream attention is then computed as
\[
\operatorname{Attn}
\left(
Q^{(\ell)}_p,\,
K^{(\ell)},\,
V^{(\ell)};
M^{(\ell)}
\right),
\]
where the mask \(M^{(\ell)}\) restricts attention to the allowed pattern.
The mask \(M^{(\ell)}\) preserves ordinary causal attention within each stream.
Across streams, it allows attention only between virtual tokens.
For a query at stream \(p\), position \(t\), and a key-value at stream \(r\), position \(s\), attention is allowed exactly when
\[
\begin{aligned}
& \left( p = r \;\land\; s \leq t \right)
\;\lor\;
\left( p \neq r \;\land\; t \in C^{(\ell)}_p \right. \\
& \left. \;\land\; s \in C^{(\ell)}_r \;\land\; s \leq t \right)
\end{aligned}
\]
Thus non-communication tokens cannot directly read another stream. Cross-stream information must be written into a communication position and then propagated forward through ordinary causal attention.

\noindent We experimented with three ways to define \(C^{(\ell)}_p\).
One approach inserts \(K\) learned vocabulary tokens (\texttt{[CROSS\_ATTEND]}) at uniform spacing through the sequence. These tokens have no loss and act only as buffers or scratchpads for moving information from one sequence to another.
Another approach keeps sequence length fixed and pre-selects \(K\) ordinary token positions, spread uniformly, as communication positions via the mask. Here the virtual role is assigned to content-bearing hidden states rather than separate buffer tokens, much as Longformer designates selected positions as global attention sites \citep{beltagy2020longformer}.

Our third approach also routes over existing positions, but learns which positions to expose with a per-layer router in the style of Mixture-of-Depths \citep{raposo2024mixtureofdepthsdynamicallyallocatingcompute}.
At each layer, a gate head scores every position,
\[
g^{(\ell)}_{p,t} = u_\ell\!\left(H^{(\ell)}_{p,t}\right),
\qquad
\pi^{(\ell)}_{p,t} = \sigma\!\left(g^{(\ell)}_{p,t}\right),
\]
where \(\sigma\) is the logistic function and \(\pi^{(\ell)}_{p,t}\) is a per-position Bernoulli parameter.
During training and prefill, we do not sample from this distribution.
Instead, we select a fixed budget of \(K\) positions per stream by top-\(K\) on the gate logits,
\[
C^{(\ell)}_p
=
\operatorname{TopK}_{t}
\left(
g^{(\ell)}_{p,t}, K
\right),
\]
which enforces a sparse cross-stream channel while keeping selection deterministic.
The sigmoid gate \(\pi^{(\ell)}_{p,t}\) then scales the attention output at selected positions, letting gradients propagate back to the router.
Because top-\(K\) routing is undefined at decode time, where only one new position is visible per step, we add an auxiliary BCE head \(v_\ell\) that predicts membership from \(H^{(\ell)}_{p,t}\) alone, enabling causal routing without revisiting past positions.
Cross-stream queries and keys are restricted to these routed sites via a FlexAttention block mask \citep{attentiongym2024, dong2024flexattentionprogrammingmodel}, which compiles to a fused Triton kernel; unless noted otherwise, our main results use this learned router, denoted \texttt{Dynamic}.

\section{Toy Tasks}

\subsection{Parallel Needle-in-a-Haystack}
\label{sec:pniah}
\noindent Needle-in-a-haystack is a popular benchmark for measuring whether language models can retrieve information from long contexts \citep{kamradt2023needle,liu2024lostmiddle,hsieh2024ruler}.
To isolate cross-stream retrieval, we introduce a synthetic benchmark called Parallel Needle-in-a-Haystack (PNIAH).
Each training example is a single long sequence partitioned into \(B\) parallel streams.
We call the stream that contains the needle the \emph{holder}. The \emph{holder} can answer the question using only local retrieval.
The remaining streams are \emph{non-holders} and their answer can only depend on information from other streams.

\noindent
We report holder accuracy \(A_H\) and non-holder accuracy \(A_{NH}\) separately. Each plotted point uses 1000 evaluation examples, so \(A_H\) is measured over 1000 holder streams and \(A_{NH}\) over \(1000(B-1)\) non-holder streams.

\begin{figure}[t]
\centering
\includegraphics[width=\columnwidth]{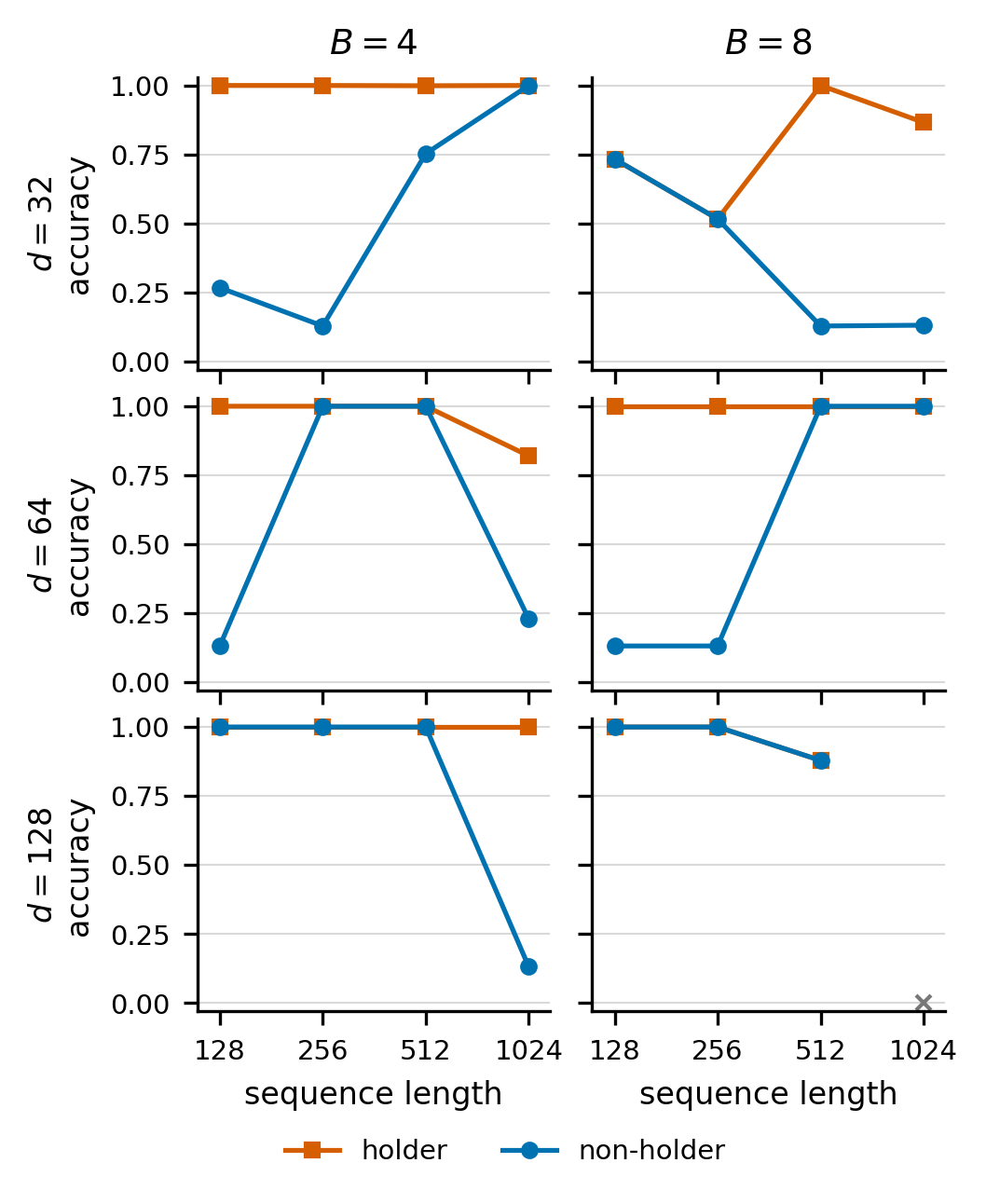}
\caption{PNIAH holder and non-holder accuracy for \(B=4\) and \(B=8\). Rows correspond to hidden sizes 32, 64, and 128, columns correspond to \(B=4\) and \(B=8\), and the \(x\)-axis is log-spaced haystack length. The gray \(\times\) marks a configuration that did not complete.}
\label{fig:niah-main-b4-b8}
\end{figure}

\noindent We conduct a sweep over sequence lengths from \(128, 256, \dots 1024\), batch sizes \(B = 4, 8\) and hidden sizes \(d = 32, 64, 128\). We use four-layer Transformer models that follow the Qwen3 architecture except for the attention mechanism \citep{yang2025qwen3}. Figure~\ref{fig:niah-main-b4-b8} visualizes holder and non-holder accuracies.
The \(x\)-axis is haystack length and the \(y\)-axis is accuracy.
Rows fix hidden size and columns fix \(B\).

\noindent Most models in this sweep solve the holder task, and larger models often solve the non-holder task as well. Counterintuitively, in some settings, non-holder accuracy increases with sequence length.

\section{Pretraining}
\label{sec:pretraining}

\noindent Having shown in toy experiments that \xattn{} can move information across sequences, we next ask whether cross-stream communication helps ordinary language-model pretraining.

\noindent For our experiments, we use TinyStories \citep{eldan2023tinystoriessmalllanguagemodels}, which is small enough for controlled architecture comparisons while still requiring coherent next-token prediction.

\noindent We compare 3 model families on TinyStories.
As a baseline reference we train a standard Qwen3 architecture \citep{yang2025qwen3}.
We train ParScale at \(B=4\) and \(B=8\) \citep{chen2025parscale} so we can compare directly against \method{} under the same multi-stream setup.
Our main models are \method{} with Static and Dynamic, at \(B=4\) and \(B=8\). 

To facilitate the learning of cross-sequence communication, we add noise in the form of masking tokens to each sequence independently, thus forcing sequences to attend to each other to recover lost information.
However for a 1-1 comparison to \parscale{} we also include training runs with masking disabled.

All models are GPT-2-scale with hyperparameters: 12 layers, \(d_{\mathrm{model}}=768\), 12 attention heads, MLP width 3072, and a 50{,}257-token vocabulary, otherwise using the Qwen3 architecture \citep{yang2025qwen3}.

\noindent In addition to next-token cross-entropy, we evaluate whether \method{} can move specific information across streams with 
a benchmark we name Parallel-QA (PQA).

Each example contains a source record, a masked query, and five candidate completions; we place the record and query in different streams.
For each candidate \(c_i=(y_1,\ldots,y_T)\), we score the teacher-forced completion as
\[
s_i = \sum_{t=1}^{T}\log p_\theta(y_t \mid x, y_{<t}).
\]
The predicted answer is the highest-scoring candidate.
Accuracy (ArgMax) reports whether that candidate is correct, while rank reports the position of the correct candidate after sorting by score. The concatenated variant puts the record and query in a single ordinary sequence, allowing us to test whether standard within-stream attention can solve the candidate-scoring problem. This setup isolates model capability from cross-stream attention, making it possible to distinguish limitations due to the model itself from those arising specifically from cross-stream communication. Appendix~\ref{app:name-probe-example} gives a concrete example of the prompt and stream layout.

\begin{figure}[tb]
\centering
\includegraphics[width=\columnwidth]{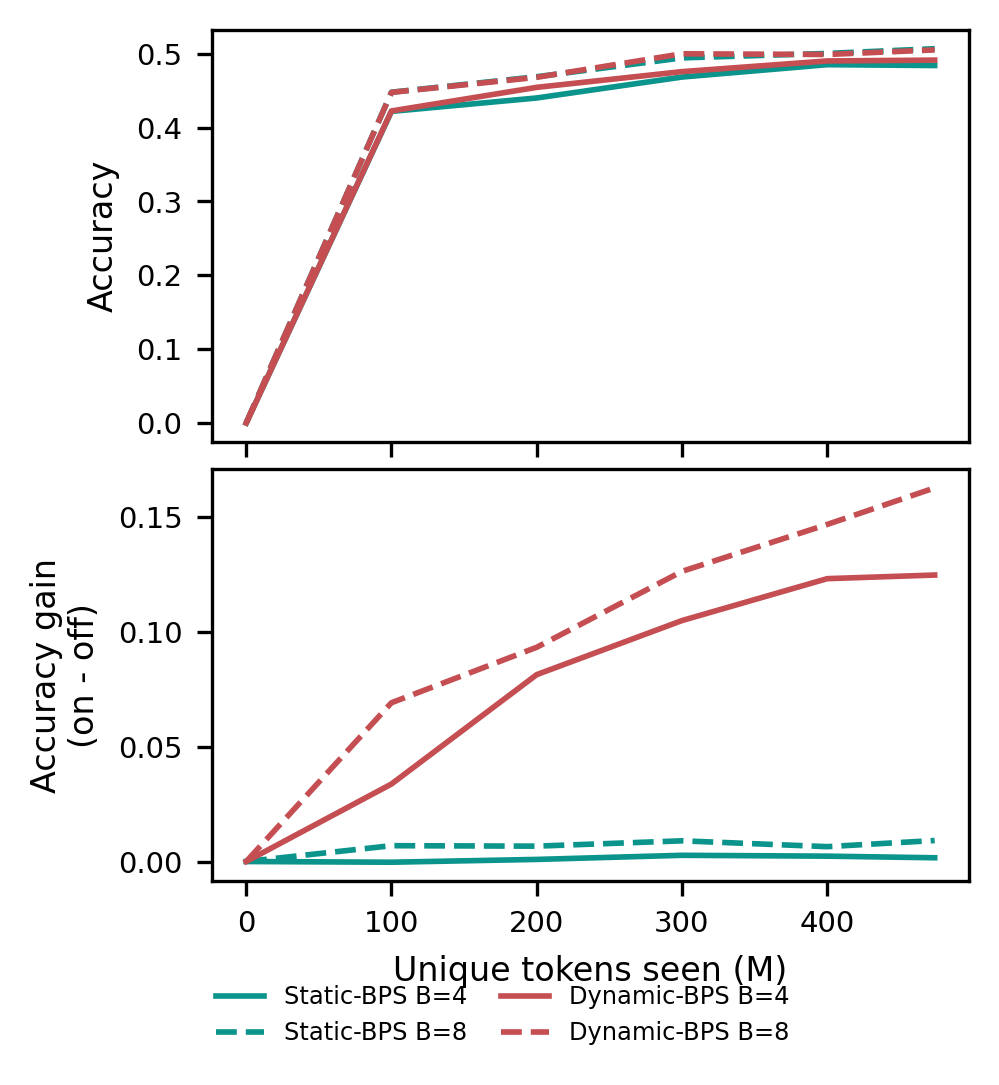}
\caption{Masked-context recovery on TinyStories. Top: masked-token argmax accuracy with cross-stream attention enabled. Bottom: accuracy gain over the same diagnostic with cross-stream attention disabled.}
\label{fig:tinystories-masked-context}
\end{figure}

\noindent
Table~\ref{tab:tinystories-pretrain-results} summarizes the final eval losses. The training-loss curves are shown in Appendix~\ref{app:tinystories-training-loss}, Figure~\ref{fig:tinystories-training-loss}.

\noindent We find that the \method{} models cluster around eval loss 1.008-1.014, a small 0.009-0.015 nats below ParScale \(B=4\).

The \(30\%\) masked variants also remain below the Qwen3 baseline \citep{yang2025qwen3}, meaning that even when a meaningful portion of the sequence is garbled, our model is able to predict the next token better than the baseline, by retrieving the right context from other sequences.
For the ParScale baselines, we use the same aggregation label-smoothing parameter as the ParScale paper, \(\epsilon=0.1\) \citep{chen2025parscale}. Even with this matched regularizer, ParScale \(B=8\) is substantially worse than ParScale \(B=4\), suggesting stream collapse or under-utilization of streams.

\begin{table}[t]
\centering
\scriptsize
\setlength{\tabcolsep}{1pt}
\begin{tabular*}{\columnwidth}{@{\extracolsep{\fill}}llrrrrr@{}}
\toprule
Model & Mask & \(B\) & Eval loss & PQA acc. & PQA rank & PQA (Concat) acc. \\
\midrule
Qwen3 baseline & - & - & 1.711 & - & - & - \\
ParScale & - & 4 & 1.023 & - & - & - \\
ParScale & - & 8 & 1.268 & - & - & - \\
Static-BPS & 0\% & 4 & 1.014 & 0.375 & 2.81 & \textbf{1.000} \\
Static-BPS & 30\% & 4 & 1.485 & 0.250 & 2.75 & \textbf{1.000} \\
Dynamic-BPS & 0\% & 4 & 1.010 & 0.250 & 3.00 & 0.250 \\
Dynamic-BPS & 30\% & 4 & 1.470 & 0.125 & 3.22 & 0.125 \\
Static-BPS & 0\% & 8 & 1.010 & 0.500 & 2.44 & \textbf{1.000} \\
Static-BPS & 30\% & 8 & 1.476 & \textbf{0.625} & \textbf{1.69} & \textbf{1.000} \\
Dynamic-BPS & 0\% & 8 & \textbf{1.008} & 0.250 & 2.69 & 0.375 \\
Dynamic-BPS & 30\% & 8 & 1.453 & 0.125 & 3.22 & 0.250 \\
\bottomrule
\end{tabular*}
\caption{TinyStories pretraining results. Eval loss is aggregate validation loss. PQA accuracy/rank evaluate a split source-query probe with \(n=8\) examples and five candidates, so chance accuracy is \(0.2\) and chance rank is about 3. PQA Concatenated accuracy uses the same probe with source and query in one ordinary sequence.}
\label{tab:tinystories-pretrain-results}
\end{table}

\noindent The concatenated column separates model capability from cross-stream communication.
Static-BPS reaches perfect concatenated accuracy, showing that ordinary attention can solve the probe when the evidence and query are in one sequence.
Its lower cross accuracy on the other hand, shows that the model is not yet able to retrieve important facts from other sequences.
Dynamic-BPS has the best aggregate losses, but remains weak even in the concatenated condition, so its low cross accuracy should not be interpreted only as a communication failure, and although its loss is slightly better than Static-BPS, it is not able to solve these. We therefore find that aggregate loss improvements do not, by themselves, demonstrate reliable communication of specific facts between streams.

\noindent Figure~\ref{fig:tinystories-masked-context} tracks masked-token argmax accuracy during pretraining on the duplicated \(30\%\)-masked validation setup.
With cross-stream attention enabled, all four models eventually reach roughly \(50\%\) argmax accuracy on masked positions.
The lower panel isolates the cross-stream contribution by subtracting the same diagnostic with cross-stream attention disabled.
Dynamic-BPS shows a clear gain, reaching about \(12\)--\(16\) percentage points by the end of training, with the largest benefit at \(B=8\).
Static-BPS remains near zero throughout, indicating that its loss improvement in Table~\ref{tab:tinystories-pretrain-results} is not driven by recovery of masked tokens through the cross-stream channel.
These results suggest that dynamic routing learns to borrow cross-stream evidence at corrupted positions, while static routing benefits mainly through generic contextual information rather than recovering specific masked tokens. This hints that alternative data augmentation methods or loss functions might be needed to better incentivize cross-stream communication.

\noindent The comparison also shows that dynamic routing is not uniformly better: it improves masked-token recovery, but it does not perform well on PQA.

\begin{figure}[t]
\centering
\includegraphics[width=\columnwidth]{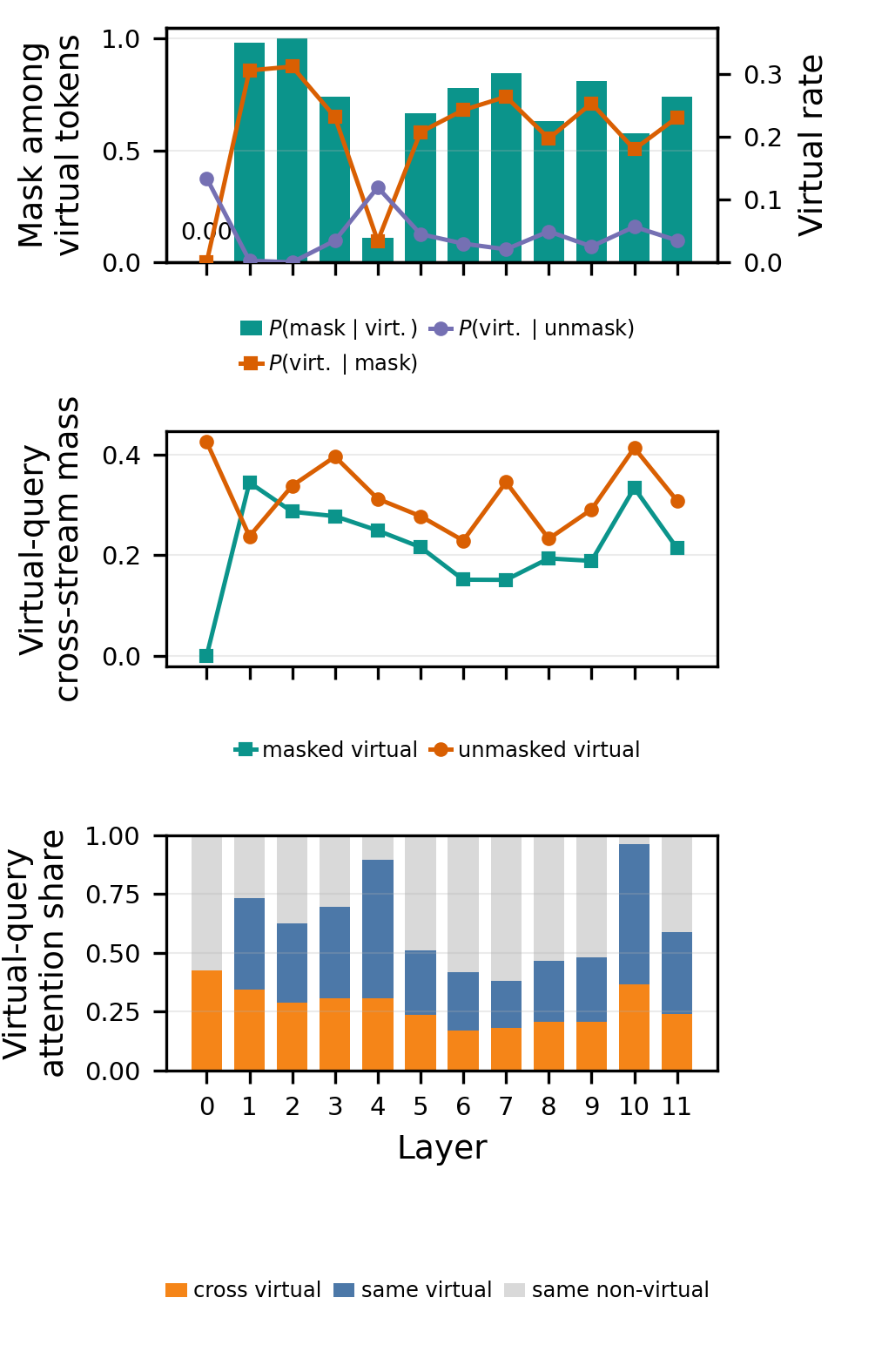}
\caption{Router diagnostics for the Dynamic-BPS \(B=4\), \(30\%\) masked TinyStories model over 500 validation chunks. Top: \(P(\mathrm{mask}\mid\mathrm{virtual})\) (bars) and \(P(\mathrm{virtual}\mid\mathrm{mask})\) / \(P(\mathrm{virtual}\mid\mathrm{unmasked})\) (lines). Middle: cross-stream attention mass from masked versus unmasked virtual queries. Bottom: full attention-budget shares for virtual queries (cross-stream virtual, same-stream virtual, same-stream non-virtual). Layer 0 routes no mask tokens as virtual.}
\label{fig:tinystories-router-diagnostics}
\end{figure}

\noindent Figure~\ref{fig:tinystories-router-diagnostics} inspects how the Dynamic-BPS router implements this behavior on 500 non-overlapping TinyStories validation chunks from the \(B=4\), \(30\%\) masked model.
The router selects 12 positions per stream in every layer.
The top panel reports both directions of mask-router coupling: the fraction of virtual tokens that are masks, and the fraction of masked or unmasked tokens that are routed as virtual.
Masks are strongly privileged as communication sites (\(P(\mathrm{virtual}\mid\mathrm{mask})\approx 0.31\) at layer 1 versus \(P(\mathrm{virtual}\mid\mathrm{unmasked})\approx 0.003\)), while layers 1 and 2 also route masks for \(98.0\%\) and \(100.0\%\) of virtual positions.
The middle panel reports mean cross-stream attention mass from virtual communication positions to virtual keys in other streams, split by whether the query token is masked or unmasked; at layer 1, masked virtual queries route more aggressively than unmasked ones (\(0.34\) versus \(0.24\)).
The bottom panel decomposes each virtual query's full attention budget into same-stream non-virtual context, same-stream virtual keys, and cross-stream virtual keys.

Together, Figures~\ref{fig:tinystories-masked-context} and~\ref{fig:tinystories-router-diagnostics} show that Dynamic-BPS learns to route cross-stream communication through corrupted local positions, but they do not yet identify whether the information comes from semantically matched neighboring streams.

\noindent To obtain visual intuition for this mechanism, we also created an interactive artifact for the Dynamic-BPS \(B=4\), \(30\%\) masked model. The viewer is available \href{https://dist-three-pi-9n64e6sfu2.vercel.app}{here}; it lets readers inspect token grids, cross-stream attention edges, and ablation heatmaps for the captured examples.
Qualitatively, it shows the same pattern as Figure~\ref{fig:tinystories-router-diagnostics}: early layers preferentially select masked tokens as communication sites, while at later layers, relay mass frequently flows through structural anchor points instead of directly targeting the exact visible copy of the masked token.
Appendix~\ref{app:positional-copy-shortcut} gives a targeted positional-offset diagnostic for these captures.
Across four masked TinyStories attention-flow runs, cross-stream attention often concentrates on the same sequence position in neighboring streams, and relay mass concentrates near the aligned position even when the aligned token in the other stream is also masked.
This suggests that the masked-copy setup can induce a positional addressing shortcut rather than fully content-addressed retrieval.

\subsection{Is \method{} Selective?}
\label{sec:source-mismatch-masked-eval}

\noindent The masked TinyStories results show that \xattn{} can improve language-model loss, but they do not identify what kind of cross-stream information is being used.
In the standard masked diagnostic, all four streams are drawn from TinyStories, so neighboring streams will always contain useful distributional context. We therefore evaluate a different setting where not all sequences will be informative, and study whether the virtual tokens have learned to be selective in what information they attend to. For this analysis, we use the Dynamic-BPS \(B=4\), \(30\%\) masked TinyStories model and compare masked-token loss with cross-stream attention enabled and disabled.

\noindent Each evaluation batch contains four streams.
We consider three versions.
First, all streams are drawn from a single TinyStories validation batch with \(30\%\) masking applied, denoted \(AAAA\) or \(BBBB\).
Second, streams are split between two disjoint TinyStories validation sequences, denoted \(AABB\).
Third, half the streams are TinyStories and half are uniformly random token sequences, denoted \(AAUU\) or \(BBUU\).
We found that stream order didn't matter for results, so we report the source layouts without separate order variants.
All results in Table~\ref{tab:source-mismatch-masked-eval} use 100 masked evaluation batches with masking rate \(0.3\) for Dynamic-BPS \(B=4\).

\begin{table*}[t]
\centering
\scriptsize
\setlength{\tabcolsep}{3pt}
\begin{tabular*}{\textwidth}{@{\extracolsep{\fill}}llrrrrrrr@{}}
\toprule
Layout & Source & CE on & CE off & \(\Delta\) CE & Mass R0 & Mass R1 & Mass R2 & Mass R3 \\
\midrule
\(AAAA\) & A & 2.517 & 3.327 & -0.811 & 0.248 & 0.258 & 0.252 & 0.241 \\
\(AABB\) & A & 2.557 & 3.311 & -0.754 & 0.366 & 0.368 & 0.134 & 0.131 \\
\(AABB\) & B & 2.711 & 3.487 & -0.777 & 0.125 & 0.128 & 0.375 & 0.372 \\
\(BBBB\) & B & 2.743 & 3.444 & -0.700 & 0.248 & 0.244 & 0.256 & 0.253 \\
\(AAUU\) & A & 2.507 & 3.319 & -0.812 & 0.367 & 0.366 & 0.132 & 0.135 \\
\(AAUU\) & U & 12.917 & 12.739 & 0.178 & 0.116 & 0.113 & 0.387 & 0.384 \\
\(BBUU\) & B & 2.695 & 3.462 & -0.766 & 0.373 & 0.355 & 0.135 & 0.137 \\
\(BBUU\) & U & 12.863 & 12.704 & 0.158 & 0.113 & 0.109 & 0.386 & 0.392 \\
\bottomrule
\end{tabular*}
\caption{Source-mismatch masked evaluation for the Dynamic-BPS \(B=4\), \(30\%\) masked TinyStories model over 100 validation batches. Row indices follow the layout string left to right (e.g.\ \texttt{AABB} assigns rows 0--1 to stream A and rows 2--3 to stream B; \texttt{AAUU} assigns rows 0--1 to A and rows 2--3 to U). Mass R\(i\) is the fraction of attention from masked router-selected queries in the source stream to router-selected keys in batch row \(i\), renormalized over all router-selected keys in the batch.}
\label{tab:source-mismatch-masked-eval}
\end{table*}

\noindent \Xattn{} improves masked CE on real TinyStories streams in every condition.
The improvement is similar when all neighboring streams are real TinyStories, when the batch mixes two disjoint TinyStories streams, and when half of the neighboring streams are random tokens.
In contrast, the random streams do not benefit.
Their loss slightly increases when cross-stream attention is enabled.

\noindent We also measure how attention is distributed over sequences.
Let \(S\) denote the set of virtual token positions in a batch and let \(\alpha_{q,k}\) be the attention weight from a virtual token \(q\) to key \(k\in S\).
We renormalize over all virtual tokens,
\[
\tilde{\alpha}_{q,k}=\frac{\alpha_{q,k}}{\sum_{k'\in S}\alpha_{q,k'}},
\]
and report the mass landing on each sequence separately (Mass R0--R3 in Table~\ref{tab:source-mismatch-masked-eval}).

\noindent We find that sequences allocate relatively little attention mass to sequences different from the source. For example, in AABB and AAUU all rows for B and U get around 0.125 of the probability mass, despite B being more informative than the random noise in U. Thus we find that the model does exhibit selectivity: it can ignore foreign sequences, but it does not distinguish between sequences that are different but still informative (B) and completely random sequences (U).

Even with noisy neighboring streams, masked CE on \(A\) or \(B\) is lower than the ordinary unmasked next-token CE of the pretrained TinyStories Qwen baseline measured on the same source-mismatch validation setup.

\section{Continued Pretraining with LoRA}
\label{sec:cpt}

\noindent We next ask whether the same mechanisms can be used when starting from an existing pretrained language model. To study this, we ran a 100M-token FineWeb-Edu continued-pretraining comparison using Qwen3-0.6B with sequence length 512 \citep{penedo2024fineweb}.
The multi-stream models use \(B=4\), \(K=12\) virtual tokens, and LoRA rank 64 with alpha 128. Table~\ref{tab:fineweb-finetune-results} reports the final LM losses and evaluation scores across PNIAH and PMHQA-style datasets. For all candidate-scoring probes, including WikiHop, we rank candidates by mean token log-probability. Training curves for these runs are shown in Appendix~\ref{app:fineweb-finetune-training-loss}, Figure~\ref{fig:fineweb-finetune-training-loss}.

\noindent As a stronger task-specific control, we also fine-tune a \method{} \(B=4\) adapter directly on WikiHop with split evidence across parallel streams (Appendix~\ref{app:additional-qa-finetuning}).
Each example forms one batch-parallel forward pass: three support documents occupy separate streams and a fourth target stream contains the query and supervised answer tokens, with LoRA on Qwen3-0.6B and dynamic cross-stream attention enabled.
Training minimizes the split-evidence objective
\[
\mathcal{L}_{\mathrm{split}}
= - \sum_{i \in \mathcal{A}_t}
\log p_\theta\!\left(y_{t,i}\mid x_{1:B}, y_{t,<i}\right),
\]
so only target answer tokens are supervised while evidence streams still influence predictions through cross-stream attention.

\begin{table*}[tb]
\centering
\scriptsize
\setlength{\tabcolsep}{2pt}
\resizebox{\textwidth}{!}{%
\begin{tabular}{llrrrrrrrrrrr}
\toprule
Model & Mask & Train & Eval & PN on & PN off & PN cat & PM on & PM off & PM cat & WH on & WH off & WH cat \\
\midrule
Qwen3 base & - & - & - & - & - & 1.000 & - & - & 1.000 & - & - & .360 \\
Qwen3 LoRA & - & 3.079 & 3.121 & - & - & 1.000 & - & - & 1.000 & - & - & .350 \\
ParScale \(B=4\) & - & 2.890 & 2.912 & - & - & - & - & - & - & - & - & - \\
\method{} \(B=4\) & 0\% & \textbf{2.889} & \textbf{2.853} & .094 & .107 & - & .260 & .350 & - & .230 & .060 & .200 \\
\method{} \(B=4\) & 30\% & 3.879 & 3.932 & \textbf{.112} & .117 & - & .220 & .340 & - & \textbf{.270} & .130 & .200 \\
\method{} \(B=4\) & 50\% & 4.682 & 4.777 & .102 & \textbf{.133} & - & \textbf{.300} & .290 & - & .210 & .110 & \textbf{.210} \\
\method{} \(B=4\) WikiHop & - & - & - & - & - & - & - & - & - & .300 & .230 & - \\
\bottomrule
\end{tabular}}
\caption{FineWeb-Edu LoRA continued-pretraining and QA evaluation results. Train and eval are ordinary FineWeb-Edu next-token losses after 100M unique training tokens. PN is Parallel-NIAH non-holder accuracy, PM is four-way Parallel-MHQA accuracy, and WH is WikiHop candidate accuracy. On/off denote cross-stream attention enabled and disabled for \method{} rows. Cat(concatenated) puts all evidence in one ordinary sequence. All candidate-scoring columns use mean token log-probability, which avoids the length bias of sequence-sum scoring. The \method{} \(B=4\) WikiHop row reports the direct split-evidence fine-tuning control from Appendix~\ref{app:additional-qa-finetuning}.}
\label{tab:fineweb-finetune-results}
\end{table*}

\noindent
The LM-loss result is encouraging for ordinary continued pretraining: ParScale and the no-mask \method{} run both improve over the Qwen3 LoRA baseline.
For the masked LoRA runs, we also log the same masked-context diagnostic used in TinyStories pretraining: masked-token argmax accuracy with cross-stream attention enabled versus disabled on duplicated validation batches.
Appendix Figure~\ref{fig:fineweb-masked-context} shows a clear gap, reaching roughly \(14\) percentage points on the \(30\%\)-mask run and \(10\) percentage points on the \(50\%\)-mask run, suggesting that the fine-tuned model uses neighboring streams to recover corrupted tokens.

\noindent The QA diagnostics in Table~\ref{tab:fineweb-finetune-results} show a sharper limitation.
The plain Qwen baselines solve the concatenated versions of PNIAH and PMHQA, so the tasks are answerable when all evidence is placed in one ordinary context.
The \method{} adapters, however, remain close to chance on non-holder PNIAH and PMHQA, and disabling cross-stream attention is often comparable or better.

WikiHop, however, shows a different pattern: the concatenated baseline reaches 36\% accuracy, whereas our CPT adapters reach 20--27\%, meaningfully above chance and close to the dedicated \method{} \(B=4\) WikiHop fine-tuning run at 30\%.

For additional notes on QA-finetuning experiments and methods see Appendix~\ref{app:additional-qa-finetuning}.

\section{Multi-Agent Reinforcement Learning}

\noindent Our longer-term motivation is not only to let parallel streams exchange information, but to let them learn coordinated strategies.
This section studies multi-agent coordination problems at a small scale.
Each agent's observation history is treated as a stream, and the policy must learn when information from other streams is useful for coordinated action.

\noindent We study six policy classes: MLP, RNN, Temporal Transformer, Agent Transformer, Temporal-Agent Transformer, and \method{}.
For each environment state, agent \(i\) at time \(t\) observes \(o_{t,i}\), which is embedded into a token \(h_{t,i}\).
The policy outputs an action distribution
\[
\pi_\theta(a_{t,i} \mid \cdot)
=
\operatorname{Softmax}(W_\pi z_{t,i}),
\]
where \(z_{t,i}\) is the final representation for that agent and timestep.
In our current implementations, transformer tokens are observation-history tokens, not separate observation and action tokens.
Previous actions are used by PPO \citep{schulman2017ppo} for the policy-gradient objective, and available-action masks are applied to the output logits, but actions are not fed back as tokens in the attention sequence.
The Temporal Transformer applies causal attention independently within each agent's history,
\[
(t,i) \leftarrow (s,i), \quad s \leq t.
\]
The Agent Transformer attends across agents only at the current timestep,
\[
(t,i) \leftarrow (t,j).
\]
The Temporal-Agent Transformer flattens time and agents into one sequence and allows dense causal attention across both axes,
\[
(t,i) \leftarrow (s,j), \quad s \leq t.
\]
At long sequences, this architecture becomes infeasible. It is also a natural upper reference for \method{}, because it can use dense attention across agents and time rather than communicating only through selected positions.

\noindent \method{} keeps a separate temporal stream for each agent and permits cross-agent communication only through selected virtual tokens,
\[
(t,i) \leftarrow (s,j)
\quad \text{only if} \quad
i \neq j,
\]
\[
t \in C_i,\quad s \in C_j,\quad s \leq t.
\]

We note that \method{} is at a significant disadvantage compared to our most expressive model Temporal-Agent-Transformer which can do full dense attention across all streams.

\noindent All policies are trained with a MAPPO-style actor-critic objective. This follows the centralized-training, decentralized-execution paradigm: the critic receives centralized world-state histories during training, while the actor acts from per-agent observation histories.
For \method{}, Agent-Transformer, and Temporal-Agent-Transformer, decentralized execution still includes learned communication, because each agent stream can read selected hidden states from other agent streams.
For each policy class, the critic receives centralized world-state histories and uses the corresponding architecture family.
The comparison therefore keeps the MAPPO-style training setup fixed across policies, but it does not hold the critic architecture fixed.

\subsection{MPE}

\noindent The MPE tasks are small but communication-sensitive coordination problems \citep{lowe2017maddpg}.
Table~\ref{tab:mpe} reports the three-seed results.
On Simple Spread and Simple Reference, \method{} is competitive with the Temporal-Agent Transformer.

\begin{table}[tb]
\centering
\footnotesize
\setlength{\tabcolsep}{2pt}
\begin{tabular*}{\columnwidth}{@{\extracolsep{\fill}}llrrrr@{}}
\toprule
Task & Policy & Final & Std. & Best & Last 10 \\
\midrule
Spread & MLP & -22.7 & 0.4 & -22.3 & -22.7 \\
Spread & RNN & -21.0 & 0.2 & -20.7 & -21.3 \\
Spread & Agent & -19.3 & 0.4 & -18.8 & -19.2 \\
Spread & Temporal & -19.2 & 0.2 & -18.8 & -19.2 \\
Spread & Temporal-Agent & -19.1 & 0.4 & \textbf{-18.3} & \textbf{-18.8} \\
Spread & \method{} & \textbf{-18.9} & 0.3 & -18.5 & -19.0 \\
\midrule
Reference & MLP & -27.0 & 0.4 & -25.8 & -26.9 \\
Reference & RNN & -21.3 & 0.3 & -20.5 & -21.2 \\
Reference & Agent & -20.6 & 0.2 & -20.3 & -21.0 \\
Reference & Temporal & -19.7 & 0.4 & -19.3 & -20.0 \\
Reference & Temporal-Agent & -20.0 & 0.5 & -19.7 & -20.3 \\
Reference & \method{} & \textbf{-19.6} & 0.3 & \textbf{-19.2} & \textbf{-19.8} \\
\bottomrule
\end{tabular*}
\caption{MPE Simple Spread and Simple Reference results, 500K timesteps per run, seeds 0--2. Higher return is better. RNN is the official recurrent baseline; Agent, Temporal, and Temporal-Agent are transformer policies.}
\label{tab:mpe}
\end{table}

\subsection{SMAX}

\noindent SMAX is a harder stress test for sparse virtual-token communication, implemented through the JaxMARL SMAX suite \citep{samvelyan2019smac,rutherford2023jaxmarl}.
Table~\ref{tab:smax} summarizes the main SMAX comparison.
Across three rerun seeds, the strongest policies vary substantially, while the temporal-attention policies fail on this task with consistently low scores.
\method{} remains close to the temporal and Temporal-Agent transformer means, but all agent communication mechanisms are weak on this task and a simple MLP performs best, suggesting multi-agent coordination is not critical for this task.

\begin{table}[tb]
\centering
\footnotesize
\setlength{\tabcolsep}{2pt}
\begin{tabular*}{\columnwidth}{@{\extracolsep{\fill}}lrrrr@{}}
\toprule
Policy & Return & Std. & Win rate & Time \\
\midrule
MLP & \textbf{1.5} & 0.7 & \textbf{0.65} & 212s \\
RNN & 1.2 & 0.6 & 0.35 & 342s \\
Agent Transformer & 1.1 & 0.5 & 0.30 & 223s \\
Temporal Transformer & 0.8 & 0.0 & 0.09 & 645s \\
Temporal-Agent Transformer & 0.8 & 0.1 & 0.09 & 1231s \\
\method{} & 0.8 & 0.0 & 0.07 & 1307s \\
\bottomrule
\end{tabular*}
\caption{JaxMARL SMAX \texttt{5m\_vs\_6m}, seeds 0--2, about 10M environment steps per seed. Higher return and win rate are better. Time is mean wall-clock time per seed. Win-rate spread was not logged for this summary.}
\label{tab:smax}
\end{table}

\subsection{MABrax}

\begin{table}[tb]
\centering
\footnotesize
\setlength{\tabcolsep}{2pt}
\begin{tabular*}{\columnwidth}{@{\extracolsep{\fill}}lrrrr@{}}
\toprule
Policy & Final & Std. & Best & Last 10 \\
\midrule
MLP & 74 & 2 & 79 & 74 \\
RNN & 97 & 11 & 113 & 96 \\
Agent Transformer & \textbf{110} & 2 & \textbf{114} & \textbf{107} \\
Temporal Transformer & 95 & 5 & 101 & 92 \\
Temporal-Agent Transformer & 96 & 7 & 99 & 93 \\
\method{} & 85 & 3 & 89 & 84 \\
\bottomrule
\end{tabular*}
\caption{MABrax HalfCheetah 6x1, seeds 0--2, 500K timesteps per run. Metrics are rollout returns; higher is better.}
\label{tab:mabrax}
\end{table}

\noindent MABrax gives a continuous-control counterexample to the MPE gains \citep{freeman2021brax,rutherford2023jaxmarl}.
Table~\ref{tab:mabrax} reports the HalfCheetah results.
\method{} learns and improves over the MLP baseline, but it underperforms the temporal and agent transformers on this HalfCheetah task.

\subsection{Jumanji Search and Rescue}

\noindent SearchAndRescue tests partial observability more directly \citep{bonnet2024jumanji}.
Table~\ref{tab:jumanji-search-rescue} reports the main SearchAndRescue setting.
For results varying the number of vision rays and searchers, see Figure~\ref{fig:search-rescue-env4-sweeps}.

\begin{table}[tb]
\centering
\scriptsize
\setlength{\tabcolsep}{2pt}
\begin{tabular*}{\columnwidth}{@{\extracolsep{\fill}}lrrrr@{}}
\toprule
Policy & Found & Std. & Remaining & Time \\
\midrule
MLP & 18.4 & 0.4 & 0.08 & 44.5s \\
RNN & 18.1 & 0.9 & 0.10 & 144.0s \\
Agent Transformer & 19.2 & 0.6 & 0.04 & 67.6s \\
Temporal Transformer & 18.4 & 1.3 & 0.08 & 110.5s \\
Temporal-Agent Transformer & \textbf{19.7} & 0.2 & \textbf{0.02} & 99.3s \\
\method{} & 19.4 & 0.4 & 0.03 & 111.8s \\
\bottomrule
\end{tabular*}
\caption{Jumanji SearchAndRescue-v0, seeds 0--4, 500K timesteps per run with 4 searchers, 20 targets, time limit 200, and 64 vision rays. Higher found and lower remaining are better.}
\label{tab:jumanji-search-rescue}
\end{table}

\noindent Appendix Table~\ref{tab:jumanji-visibility} reports a sweep over the visibility parameter. \method{} beats Temporal-Agent in four of six settings.

\noindent Overall, \method{} participates in learned coordination: strong on MPE, competitive on SearchAndRescue.
They are also mixed: \method{} underperforms on MABrax and does not solve the harder SMAX setting.
Sparse communication is therefore useful, but not a complete substitute for denser temporal-agent attention.

\section{Performance and Hardware-Aware Cross-Stream Attention Mechanisms}
\label{sec:perf}

\begin{figure}[tb]
\centering
\includegraphics[width=0.92\columnwidth]{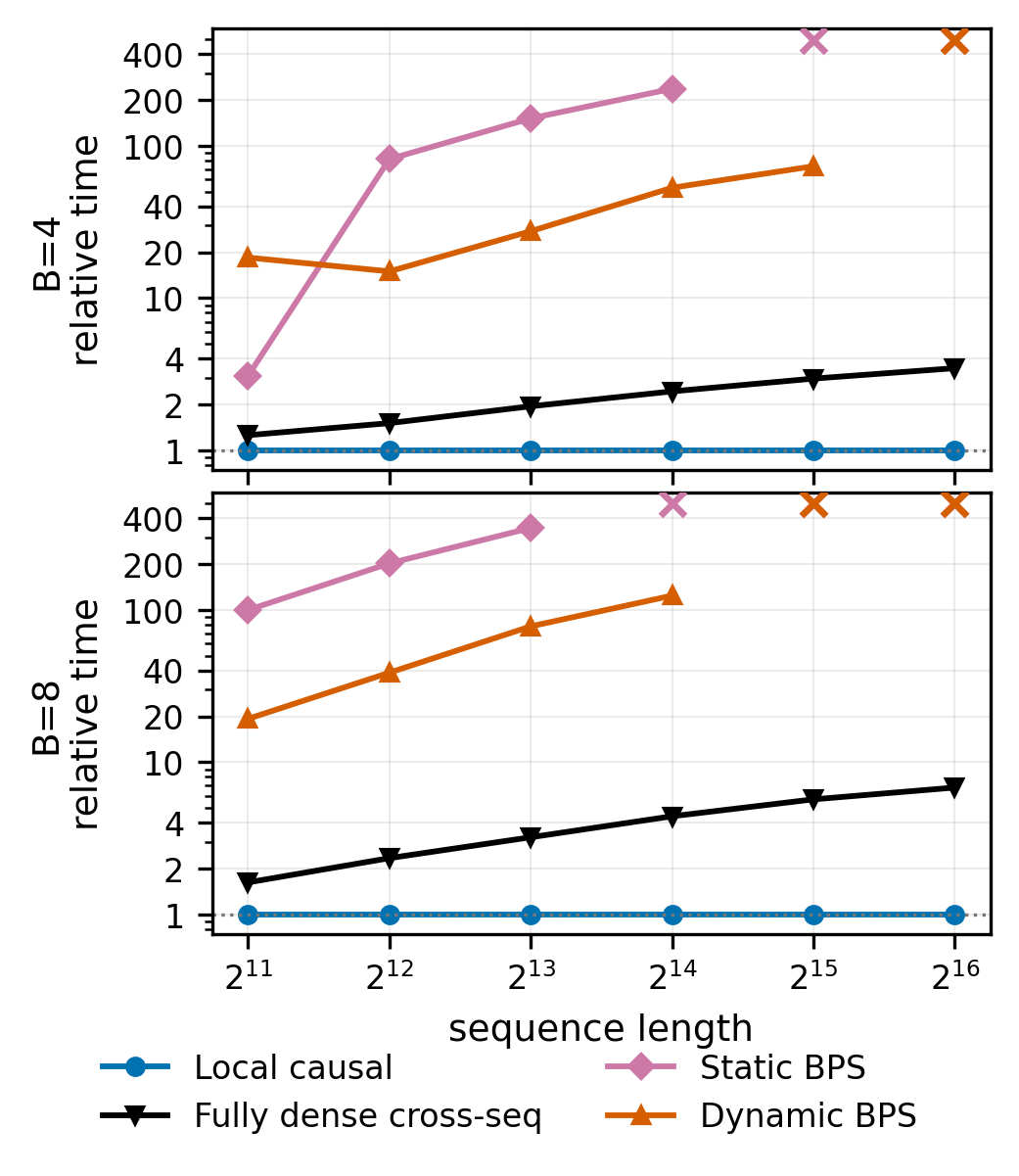}
\caption{Single-layer prefill/block overhead on an A100 for \(B=4,8\) at \(K/n=1/32\), normalized by local causal attention. Crosses mark out-of-memory runs. The direct masked implementation can erase the expected advantage over dense cross-stream attention.}
\label{fig:hardware-current-bps-prefill}
\end{figure}

\noindent The computational appeal of \(\method{}\) is that streams or sequences communicate through a small set of tokens rather than through fully dense attention.
With \(B\) streams of length \(n\), independent causal attention costs \(O(Bn^2)\), while allowing every token in every stream to attend across all streams costs \(O(B^2n^2)\).
If each stream exposes only \(K=n_{\mathrm{virtual}}\) communication positions, the intended cost becomes \(O(Bn^2 + B^2K^2)\).
However, Figure~\ref{fig:hardware-current-bps-prefill} shows that the asymptotic advantage does not show up in practice, and \method{} has a large overhead and even runs slower than fully dense cross-stream attention.

\noindent The direct implementation is slow because FlexAttention does not handle the resulting irregular mask efficiently.

We thus propose a two-pass implementation.
Let \(S_p\) be the selected positions in stream \(p\), and let \(H_p \in \mathbb{R}^{n \times d}\) be its hidden states.
The first pass runs local causal attention within each stream.
The second pass gathers the selected states into a compact block:
\[
\begin{aligned}
Z &= \operatorname{Gather}\left(\{H_p[S_p]\}_{p=1}^{B}\right), \\
Z &\in \mathbb{R}^{BK \times d}.
\end{aligned}
\]

\noindent For a selected query \(i\), let \(A_i\) be its local-causal key set and let \(G_i\) be the gathered selected-key set with keys already present in \(A_i\) removed. The two attention passes produce local and global outputs \(y_i^A\) and \(y_i^G\), while retaining their log normalizers:
\[
\begin{aligned}
s_i^A &= \log \sum_{j\in A_i}\exp(q_i^\top k_j/\sqrt{d}), \\
s_i^G &= \log \sum_{j\in G_i}\exp(q_i^\top k_j/\sqrt{d}).
\end{aligned}
\]
The virtual token output is then merged as
\[
y_i^{\mathrm{2pass}} =
\frac{\exp(s_i^A)y_i^A + \exp(s_i^G)y_i^G}{\exp(s_i^A)+\exp(s_i^G)}.
\]
Because \(A_i\) and \(G_i\) are disjoint, this is exactly the same softmax as attention over \(A_i \cup G_i\), which is the original selected-token mask. The merge is the standard log-sum-exp combination used in FlashAttention-style tiled kernels.
The cost is \(O(Bn^2)+O((BK)^2)\), and the \(O((BK)^2)\) term is computed by its own small dense attention kernel.

\noindent Another benefit of this decomposition is that, as in Longformer~\citep{beltagy2020longformer}, we can use separate query, key, and value projections for the local causal pass and for attention over gathered virtual-token states, without changing the merged output above.
Dedicated projections would let virtual-token queries and keys specialize for cross-stream retrieval rather than reusing representations tuned for local language modeling.

\begin{figure}[tb]
\centering
\includegraphics[width=0.92\columnwidth]{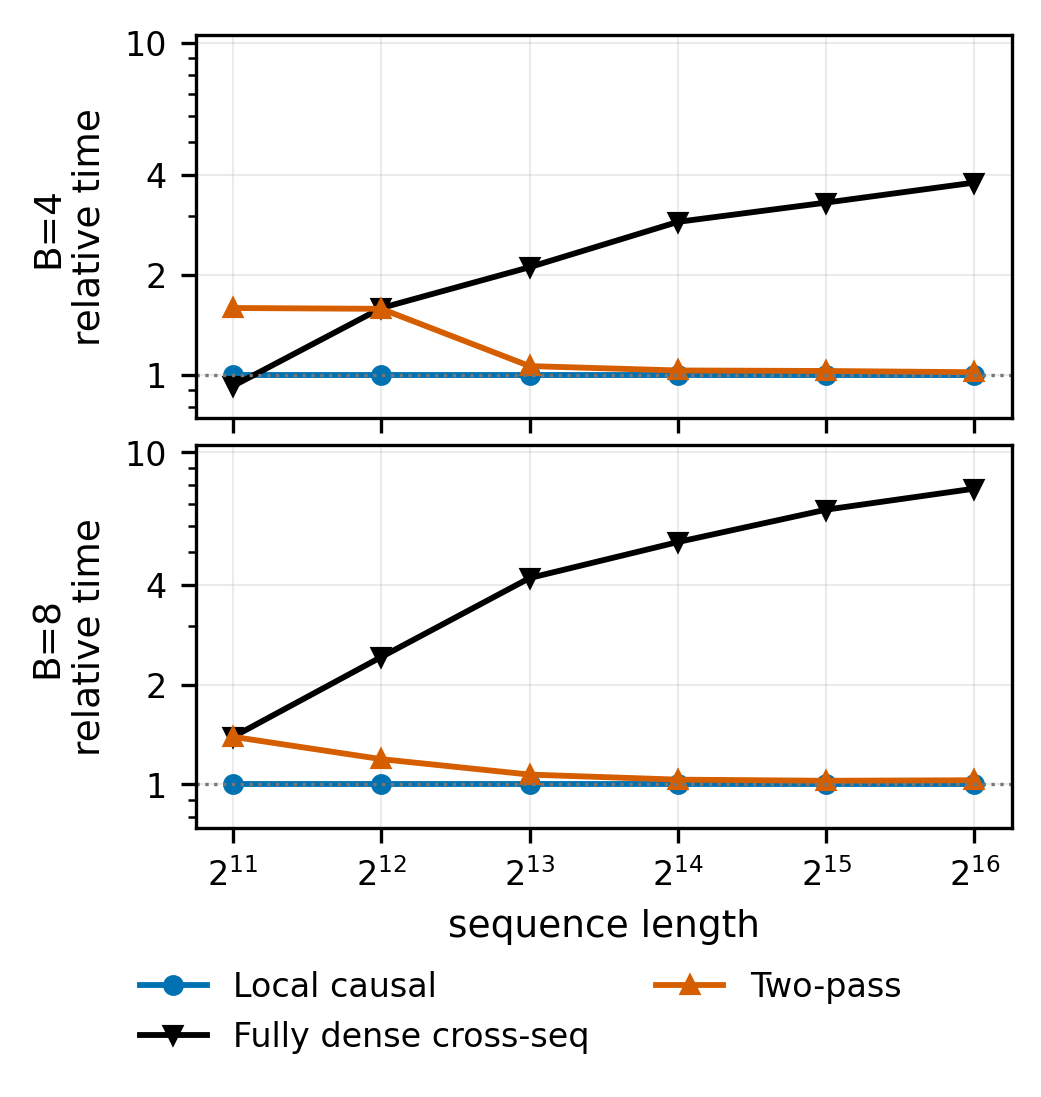}
\caption{Single-layer timing normalized by local causal attention. Dense cross-stream attention grows expensive with sequence length and batch count; exact two-pass attention stays close to local causal attention while matching the selected-token output.}
\label{fig:hardware-causal-overhead}
\end{figure}

\noindent Figure~\ref{fig:hardware-causal-overhead} compares our two-pass implementation with local causal attention and dense attention across sequences. Two-pass \method{} remains close to the local causal baseline, while fully dense attention becomes increasingly expensive as sequence length and batch count grow.

\section{Conclusion}

\noindent In this paper we introduce \xattn{} as a new class of sequence models for coordinated parallel computation. For a concrete and hardware-efficient instantiation of such a model we proposed \method{}, and showed that, when implemented carefully, it can exhibit negligible overhead compared to standard self-attention. We showed it can learn controlled cross-stream retrieval on \pniah{}, outperforms \parscale{} on language modelling on TinyStories and FineWeb-Edu, and can perform competitively on many MARL tasks.

\noindent Although 30\%-masked \method{} models beat the Qwen baseline on pretraining loss and cross-stream attention improves masked-token recovery, we find that on PNIAH all models meaningfully underperform the concatenated version, suggesting that streams still cannot reliably recover facts from parallel streams. PMHQA remains near chance despite strong holder accuracy, suggesting that language-modelling loss improvements do not yet correlate with reliable retrieval to streams that must answer without local evidence. This suggests that our current pretraining approach alone does not incentivize the cross-stream retrieval and communication these benchmarks measure. Also, whereas \method{} at 30\% noise was able to outperform the baseline on pretraining, we do not find this to be the case for finetuning. Whether this is due to the much harder language-modelling objective or the smaller space of weights to tune remains in question.

\noindent Future work should focus on better training objectives and data augmentation techniques to incentivize multi-sequence communication and coordination. For example, instead of just masking, we could mix in or rephrase data from other streams to encourage true cross-sequence reasoning instead of copying. Additionally, future experiments should further extend pretraining and finetuning to larger model sizes, helping us understand how this method performs at larger scales and on more challenging data mixtures.

\appendix

\section{SearchAndRescue Visibility Sweep}

\begin{table}[H]
\centering
\footnotesize
\setlength{\tabcolsep}{1pt}
\begin{tabular*}{\columnwidth}{@{\extracolsep{\fill}}lrrrrr@{}}
\toprule
Setting & Vision & Agent & Temporal-Agent & \method{} & \(\Delta\) \\
\midrule
Sparse & 4 & 17.29 & 16.87 & \textbf{17.18} & +0.31 \\
Sparse & 8 & 15.69 & \textbf{16.23} & 15.49 & -0.74 \\
Sparse & 16 & 15.84 & 15.52 & \textbf{16.18} & +0.66 \\
Sparse & 24 & 15.78 & 16.01 & \textbf{16.53} & +0.52 \\
Large & 12 & \textbf{12.39} & 10.55 & 11.87 & +1.32 \\
Large & 16 & 10.22 & 10.16 & \textbf{10.58} & +0.41 \\
\bottomrule
\end{tabular*}
\caption{SearchAndRescue visibility sweep, reporting final targets found over five seeds. Sparse settings use env size 2.0, 4 searchers, 25 targets, and time limit 300; large settings use env size 4.0 and time limit 800. \(\Delta\) is \method{} minus Temporal-Agent. Higher is better.}
\label{tab:jumanji-visibility}
\end{table}

\section{TinyStories Training Loss}
\label{app:tinystories-training-loss}

\begin{figure}[H]
\centering
\includegraphics[width=\columnwidth]{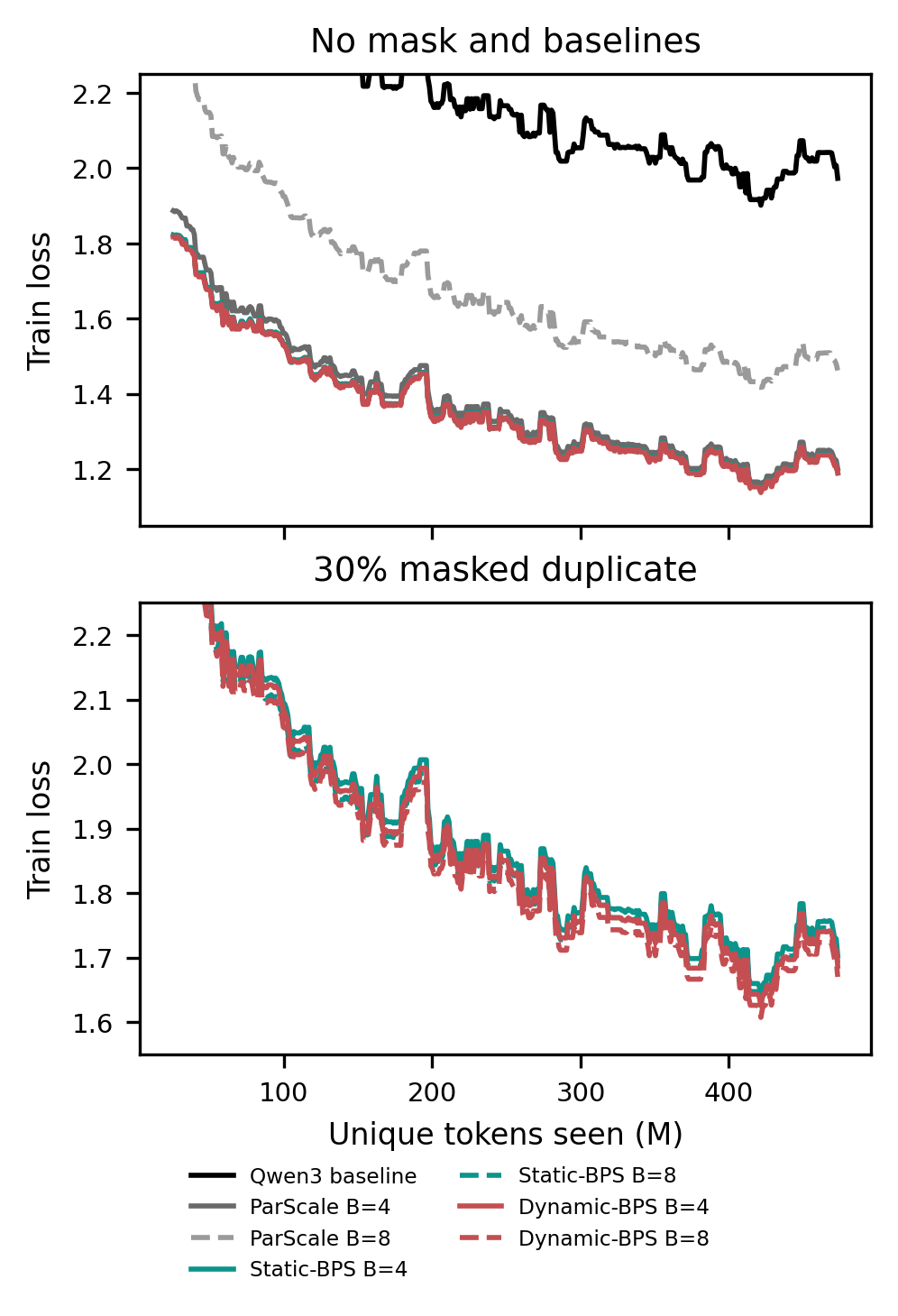}
\caption{Zoomed TinyStories training loss over unique tokens for the same runs summarized in Table~\ref{tab:tinystories-pretrain-results}. The plot omits the first 25M unique tokens and applies a 9-point median smoothing window so that late-training differences are visible. The top panel shows no-mask Static-BPS and Dynamic-BPS runs together with Qwen3 and ParScale baselines; the BPS curves cluster below ParScale \(B=4\), while the y-axis zoom clips the highest-loss baseline segments. The bottom panel shows the 30\% masked duplicate runs.}
\label{fig:tinystories-training-loss}
\end{figure}

\section{Positional Copy-Shortcut Diagnostics}
\label{app:positional-copy-shortcut}

\noindent The masked TinyStories experiments duplicate a validation sequence across parallel streams and independently mask each stream.
This setup is useful because a masked token in one stream may be visible in another stream.
However, exact duplication also creates a possible shortcut: the model can address the same sequence position in another stream rather than retrieve by content.
We therefore run a positional-offset diagnostic on four Dynamic-BPS attention-flow captures with \(B\in\{4,8\}\), masking rates \(30\%\) and \(50\%\), and sequence length 128.

\noindent For every cross-stream attention edge from query position \(t\) to key position \(s\), we record the relative offset
\[
\Delta = s - t .
\]
We aggregate attention mass over \(\Delta\) per layer and per head.
A concentration at \(\Delta=0\) means that a head reads the aligned position in another stream.
For masked queries, we additionally inspect relay sources from the attention-flow artifact.
The relay score composes head-mean local attention before and after one cross-stream bridge, then records the largest source positions contributing to each masked query.
This gives a coarse source-level view of where cross-stream information appears to originate.

\begin{table}[H]
\centering
\scriptsize
\setlength{\tabcolsep}{2pt}
\begin{tabular*}{\columnwidth}{@{\extracolsep{\fill}}lrrr@{}}
\toprule
Run & \(\Delta=0\) oracle & \(\Delta=0\) non-oracle & \(|\Delta|\leq 4\) \\
\midrule
\(B=4\), mask 30\% & 2.3 & 2.8 & 37.2 \\
\(B=4\), mask 50\% & 1.1 & 7.7 & 40.6 \\
\(B=8\), mask 30\% & 0.1 & 7.2 & 40.9 \\
\(B=8\), mask 50\% & 0.2 & 4.8 & 42.4 \\
\bottomrule
\end{tabular*}
\caption{Relay-source offset diagnostics. Values are percentages of top-source relay mass. Oracle means the aligned position in another stream contains the visible copy of the masked token. Non-oracle means the aligned position is also masked, so the token identity is not available at that location.}
\label{tab:positional-copy-shortcut}
\end{table}

\noindent The diagnostic shows two related effects.
First, some heads are strongly position locked.
In the \(B=8\) runs, late-layer heads such as L9H4, L9H7, and L9H9 put roughly \(85\%\)--\(94\%\) of their cross-stream attention mass on \(\Delta=0\).
These heads mostly read the same position in another stream.
Second, the relay-source diagnostic shows that exactly aligned relay mass is often larger for non-oracle aligned positions than for oracle aligned positions.
That is, relay mass can land on the same position in a neighboring stream even when that position is also masked.

\noindent The relay mass is not only concentrated at the exact aligned token.
Around \(37\%\)--\(42\%\) of top-source relay mass lies within four tokens of the aligned position, far above the uniform baseline of about \(3.5\%\) for a length-128 sequence.
The largest offsets are typically just before the query position.
This suggests a mechanism in which the model imports local context from the aligned neighborhood of another stream, rather than directly copying the visible target token itself.

\noindent These results support a positional-addressing interpretation of the masked-copy gains.
Cross-stream attention appears to exploit the fact that the streams are exact aligned copies, so the same position or nearby positions in another stream are likely to contain useful context.
This is useful for masked-token recovery, but it is weaker than content-addressed retrieval.
It also helps explain why aggregate masked-token improvements do not necessarily translate into reliable transfer of non-redundant facts in PNIAH, PMHQA, or WikiHop-style probes.

\noindent This diagnostic has important limitations.
It uses attention weights rather than value norms or direct logit attribution, and the relay source analysis is truncated to the top recorded sources per masked query.
The relay computation also uses head-mean attention and a restricted local-to-cross-to-local path form.
A stronger causal test would shift the duplicated streams by \(k\) tokens and measure masked-token recovery as a function of \(k\), or inject disagreeing copies and test whether predictions follow the copied stream against local context.

\section{FineWeb Continued-Pretraining Training Loss}
\label{app:fineweb-finetune-training-loss}

\begin{figure}[H]
\centering
\includegraphics[width=\columnwidth]{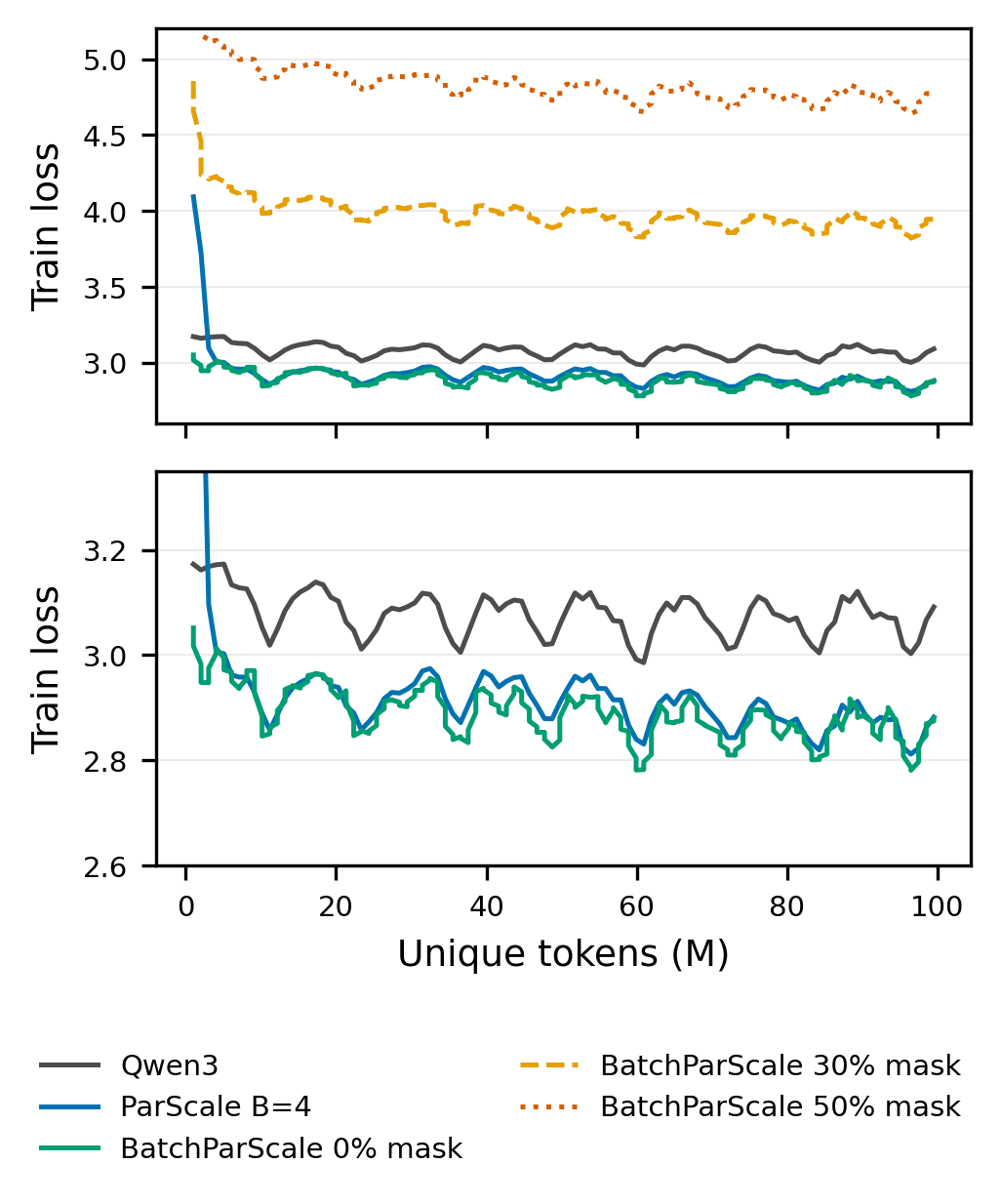}
\caption{FineWeb-Edu LoRA continued-pretraining training loss over unique tokens for the runs summarized in Table~\ref{tab:fineweb-finetune-results}. The top panel shows all five runs. The bottom panel zooms in on the Qwen3, ParScale, and no-mask \method{} runs. The masked \method{} variants have substantially higher language-model loss, even though the no-mask \method{} run is competitive with the regular ParScale baseline. Periodic oscillations reflect the fixed ordering of the training stream.}
\label{fig:fineweb-finetune-training-loss}
\end{figure}

\section{FineWeb Masked-Context Accuracy}
\label{app:fineweb-masked-context}

\begin{figure}[H]
\centering
\includegraphics[width=\columnwidth]{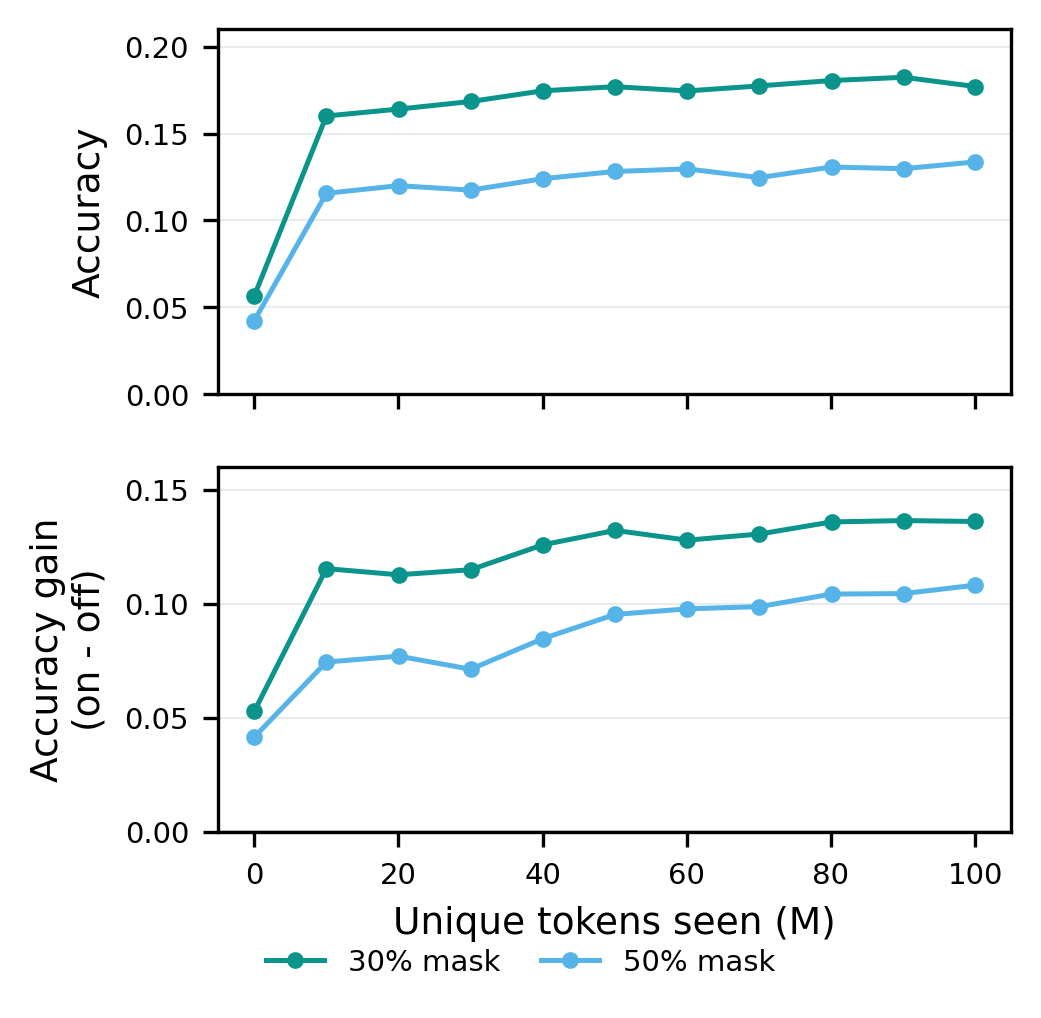}
\caption{Masked-context recovery diagnostics for the masked FineWeb-Edu LoRA continued-pretraining runs summarized in Table~\ref{tab:fineweb-finetune-results}. The top panel evaluates masked-token argmax accuracy with cross-stream attention enabled. The bottom panel reports accuracy gain over the same diagnostic with cross-stream attention disabled.}
\label{fig:fineweb-masked-context}
\end{figure}

\section{Cross-Stream Candidate Probe Example}
\label{app:name-probe-example}

\noindent The first deterministic TinyStories candidate-scoring probe example, using seed 0, is:

\begin{quote}
\small
\textbf{Source record:}
Alice and Jack went to the school yard. They met Sam near a paper crown. Sam smiled and asked their names. Alice said, ``We are Alice and Jack.''

\textbf{Masked query:}
[MASK] and [MASK] went to the school yard. They met Sam near a paper crown. Sam asked their names. They answered, ``Our names are

\textbf{Correct completion:}
Alice and Jack.

\textbf{Candidates:}
Alice and Sam.; Lily and Max.; Alice and Jack.; Jane and Kate.; Sam and Jack.
\end{quote}

\noindent The concatenated control puts the record and query in one ordinary sequence:
\begin{quote}
\small
Read the record and complete the answer.

Record: Alice and Jack went to the school yard. They met Sam near a paper crown. Sam smiled and asked their names. Alice said, ``We are Alice and Jack.''

[MASK] and [MASK] went to the school yard. They met Sam near a paper crown. Sam asked their names. They answered, ``Our names are \(\langle\)candidate\(\rangle\)
\end{quote}

\noindent The cross-stream condition for \(B=4\) separates the evidence from the target stream:
Table~\ref{tab:name-probe-stream-layout} gives the stream layout.

\begin{table}[H]
\centering
\footnotesize
\setlength{\tabcolsep}{3pt}
\begin{tabular*}{\columnwidth}{@{\extracolsep{\fill}}ll@{}}
\toprule
Stream & Text used while scoring one candidate \\
\midrule
0 & Source record only \\
1 & Masked query followed by \(\langle\)candidate\(\rangle\) \\
2 & Neutral background note about the school yard. \\
3 & Neutral background note about the paper crown. \\
\bottomrule
\end{tabular*}
\caption{Stream layout for the cross-stream candidate probe. The callback repeats this layout once for each candidate and scores the candidate tokens on stream 1.}
\label{tab:name-probe-stream-layout}
\end{table}

\noindent Thus, in the cross-stream condition, stream 1 can only prefer the correct candidate if information from stream 0 is made available through the cross-stream communication path.

\section{Multi-Stream Generation Example}
\label{app:multistream-generation-example}

\noindent Table~\ref{tab:tinystories-multistream-generation-example} shows a qualitative sample from the completed TinyStories Static-BPS \(B=4\), no-mask run. We use greedy decoding for 32 new tokens from four related toy-shop prompts. The example is qualitative only, but it illustrates the intended inference layout: several streams are decoded in parallel and each stream maintains a distinct continuation.

\begin{table*}[t]
\centering
\footnotesize
\setlength{\tabcolsep}{4pt}
\begin{tabular*}{\textwidth}{@{\extracolsep{\fill}}lp{0.32\textwidth}p{0.52\textwidth}@{}}
\toprule
Stream & Prompt & Greedy continuation \\
\midrule
0 & In the toy shop, Anna picked up a tiny drum. & She was so excited! She took it home and started to play. Anna's mom saw her playing and said, ``That drum looks so cute!'' \\
1 & In the toy shop, Ben looked for a yellow train. & He found one and said, ``I want this one!'' The shopkeeper said, ``That's a nice train. It will fit in your hand'' \\
2 & In the toy shop, the shopkeeper smiled and opened a box. & Inside the box was a toy car. The car was red and shiny. The shopkeeper said, ``This car is for you!'' The little boy \\
3 & In the toy shop, a kitten slept under the shelf. & The kitten was very cute and soft. Lily wanted to take the kitten home, but her mom said no. Lily was sad, but she listened \\
\bottomrule
\end{tabular*}
\caption{Generated multi-stream TinyStories sample from the completed Static-BPS \(B=4\), no-mask run \texttt{udilkiu0}. Continuations are greedy decodes with \texttt{max\_new\_tokens=32}.}
\label{tab:tinystories-multistream-generation-example}
\end{table*}

\noindent Table~\ref{tab:tinystories-masked-generation-example} gives a second example from the completed dynamic local-causal \(B=4\), 30\% mask run.
Here we begin from ordinary masked-evaluation copies of the same TinyStories validation sequence, then sample 20 new tokens from each masked stream without applying any further masking during generation.
The prefixes shown are the final story segment of the full 128-token prompt.

\begin{table*}[t]
\centering
\footnotesize
\setlength{\tabcolsep}{4pt}
\begin{tabular*}{\textwidth}{@{\extracolsep{\fill}}lp{0.46\textwidth}p{0.36\textwidth}@{}}
\toprule
Stream & Masked prefix tail & Sampled continuation \\
\midrule
0 & Once \(\langle\)mask\(\rangle\) a time, in \(\langle\)mask\(\rangle\) big \(\langle\)mask\(\rangle\), there lived a rhinoceros named Roxy. Roxy loved to climb. She climbed \(\langle\)mask\(\rangle\) \(\langle\)mask\(\rangle\), rocks, and hills \(\langle\)mask\(\rangle\) One day, Roxy found an icy \(\langle\)mask\(\rangle\) & . She thought it would be fun to climb it. Roxy climbed and climbed, but \\
1 & Once upon a \(\langle\)mask\(\rangle\), in a big forest, \(\langle\)mask\(\rangle\) lived a rhinoceros \(\langle\)mask\(\rangle\) R \(\langle\)mask\(\rangle\). R \(\langle\)mask\(\rangle\) loved \(\langle\)mask\(\rangle\) climb. She climbed trees, \(\langle\)mask\(\rangle\) \(\langle\)mask\(\rangle\), and hills. One \(\langle\)mask\(\rangle\), R \(\langle\)mask\(\rangle\) found an icy hill & . It was so cold, she thought she would freeze! Remy wanted to see how \\
2 & Once upon \(\langle\)mask\(\rangle\) time, \(\langle\)mask\(\rangle\) a big forest \(\langle\)mask\(\rangle\), there \(\langle\)mask\(\rangle\) a rhin \(\langle\)mask\(\rangle\) eros named \(\langle\)mask\(\rangle\) oxy. Roxy loved to climb. She climbed trees, rocks, \(\langle\)mask\(\rangle\) \(\langle\)mask\(\rangle\). \(\langle\)mask\(\rangle\) day, Roxy found an icy hill & . She thought it would be fun to climb the icy hill. Roxy started to climb \\
3 & \(\langle\)mask\(\rangle\) upon a time \(\langle\)mask\(\rangle\), in a \(\langle\)mask\(\rangle\) forest, there lived \(\langle\)mask\(\rangle\) \(\langle\)mask\(\rangle\) \(\langle\)mask\(\rangle\) oc \(\langle\)mask\(\rangle\) named Roxy \(\langle\)mask\(\rangle\) Roxy \(\langle\)mask\(\rangle\) to \(\langle\)mask\(\rangle\) \(\langle\)mask\(\rangle\) \(\langle\)mask\(\rangle\) \(\langle\)mask\(\rangle\) trees, rocks, and hills. One day \(\langle\)mask\(\rangle\), Roxy \(\langle\)mask\(\rangle\) \(\langle\)mask\(\rangle\) \(\langle\)mask\(\rangle\) hill & and decided to climb it. As Roxy climbed, she met a tiny bird named Ch \\
\bottomrule
\end{tabular*}
\caption{Masked-prefix multi-stream TinyStories sample from the completed dynamic local-causal \(B=4\), 30\% mask Hugging Face model. Each row is one masked copy of the same validation prefix.}
\label{tab:tinystories-masked-generation-example}
\end{table*}

\noindent The second stream invents the name ``Remy'' even though the unmasked story is about Roxy, illustrating that masked-prefix recovery remains imperfect.

\section{Parallel-MHQA Fine-Tuning}
\label{app:additional-qa-finetuning}

\noindent This appendix gives the details behind the QA diagnostics summarized in Table~\ref{tab:fineweb-finetune-results}.
We use these probes to test whether the masked-context benefit in continued pretraining corresponds to usable cross-stream transfer.

\noindent Parallel-NIAH (PNIAH) extends the holder/non-holder setup from Section~\ref{sec:pniah} to real FineWeb-Edu documents.
Each example has \(B=4\) parallel streams, one randomly chosen \emph{holder} stream contains a synthetic needle of the form \texttt{The secret access code is ``XXX-YYYY''}, and the remaining \emph{non-holder} streams contain unrelated text with the answer redacted.
At evaluation time, every stream is asked for the code and scored with a 10-way multiple-choice test over synthetic codes in the same format: the true answer plus nine randomly generated distractors.
For each candidate we teacher-force the completion and rank options by mean token log-probability, following the same candidate-scoring procedure as ParallelQA; chance is \(0.1\).
We report non-holder accuracy \(A_{NH}\), i.e.\ whether streams that do not locally contain the needle select the correct code.
Holder accuracy measures only local retrieval and is therefore not the metric of interest here.
This design avoids treating target-row shortcuts or memorized NIAH benchmark facts as evidence of cross-stream communication.

\noindent Parallel-MHQA (PMHQA) extends ParallelQA to a split-evidence multi-hop setting.
ParallelQA in Section~\ref{sec:pretraining} tests single-hop cross-stream retrieval: a source record in one stream must help a masked query in another select among five name completions.
PMHQA is not built from an existing multi-hop benchmark; we generate 100 fully synthetic two-hop cases with random person names, project titles, and reference codes.
Each case is authored as two short records that form a controlled chain; for example, ``Record 1: Aria assigned reference code ABC-123 to the Cedar initiative'' and ``Record 2: Milo says Nora should use Aria's reference for the Cedar initiative''.
We place the two records on separate streams while a target stream asks which code Nora should use.
The target stream lists four synthetic code options and is scored with the same teacher-forced candidate-ranking procedure as ParallelQA; chance and the best constant-position baseline are both \(0.25\).
Answering therefore depends on chaining both facts through cross-stream attention, not on retrieving a single locally absent fact or on pretrained WikiHop-style knowledge.
We run this probe on continued-pretraining adapters rather than the TinyStories from-scratch models, because the multi-hop composition is substantially harder and would be difficult to interpret on models whose reasoning capacity is still largely untrained.

\noindent Both communication diagnostics remain weak.
On PNIAH, \(A_{NH}\) remains close to chance even though holder accuracy is near perfect.
The multi-hop QA probe is also near chance: the \method{} runs reach \(0.26\), \(0.22\), and \(0.30\), and the controls with cross-stream attention disabled are \(0.35\), \(0.34\), and \(0.29\), respectively.
On two of the three settings, disabling the cross-stream channel is better, so the channel can actively hurt this probe or the differences are noise at \(n=100\).
For this task we also trained a non-holder-only objective,
\[
\mathcal{L}_{\mathrm{nonholder}}
= - \sum_{b \ne h}\sum_{i \in \mathcal{A}_b}
\log p_\theta\!\left(y_{b,i}\mid x_{1:B}, y_{b,<i}\right),
\]
where \(h\) is the holder stream.
Fine-tuning an existing LM can preserve or mobilize local knowledge and can lower language-model loss, but reliable transfer to streams that do not contain the evidence remains unresolved.
Reliable transfer to non-holder streams is the clearest failure we observe, and it makes routing the natural target for future work.

\noindent We also run direct split-evidence QA fine-tuning probes.
The goal is to test a stronger intervention than continued pretraining: if we explicitly fine-tune the model on answer supervision where evidence is placed in neighboring streams, does it learn to retrieve those facts through \xattn{}?
Table~\ref{tab:qa-finetuning-master} summarizes these direct fine-tuning controls.
These experiments are useful mainly as controls, because standard QA datasets can be answered from target-row artifacts or pretrained knowledge.
Each example is represented as \(B\) streams.
Some streams contain evidence, while one target stream contains the question and answer format.
Unless otherwise stated, only answer tokens on the target stream contribute to the supervised loss:
\[
\mathcal{L}_{\mathrm{split}}
= - \sum_{i \in \mathcal{A}_t}
\log p_\theta\!\left(y_{t,i}\mid x_{1:B}, y_{t,<i}\right),
\]
where \(\mathcal{A}_t\) is the set of answer-token positions in target stream \(t\).
The evidence streams still participate in the forward pass and can influence the target prediction through cross-stream attention, but they are not asked to reproduce the target answer.
For the matched isolated control we disable \xattn{} and train the same target loss,
\[
\mathcal{L}_{\mathrm{isolated}}
= - \sum_{i \in \mathcal{A}_t}
\log p_\theta\!\left(y_{t,i}\mid x_t, y_{t,<i}\right).
\]
A high isolated-control result means that the benchmark is not cleanly measuring cross-stream evidence.
It suggests that pretrained knowledge, benchmark artifacts, or target-row shortcuts can explain the apparent gain.

\begin{table}[H]
\centering
\footnotesize
\setlength{\tabcolsep}{4pt}
\begin{tabular*}{\columnwidth}{@{\extracolsep{\fill}}lcccc@{}}
\toprule
Task & Train & Eval & Ctrl. & Cross \\
\midrule
WikiHop & 800 & 100 & .230 & .300 \\
QASC & 800 & 926 & .454/.185 & .472 \\
Guarded QA & 800 & 926 & .125 & .146 \\
QA-v4 & 800 & 128 & .117 & .164 \\
\bottomrule
\end{tabular*}
\caption{Direct split-evidence QA fine-tuning probes. Values are accuracies. Control is isolated split accuracy, except QASC reports question/options-only followed by isolated split accuracy. Cross-stream is the best split-evidence fine-tuning result. WikiHop uses candidate scoring with chance about \(0.047\); the other multiple-choice tasks have chance \(0.125\). QA-v4 has 926 validation examples; for comparability across these runs we report the fixed 128-example evaluation.}
\label{tab:qa-finetuning-master}
\end{table}

\noindent We ran these probes on WikiHop-style candidate scoring \citep{welbl2018constructing}, QASC \citep{khot2020qasc}, and two guarded synthetic multiple-choice tasks.
WikiHop reaches 0.300 in the best direct split-evidence run, but the isolated split control already reaches 0.230, far above the random-candidate baseline.
QASC is even less clean: a base model with only the question and answer options reaches 0.454, close to the best split cross-stream result of 0.472.
These results show that pretrained knowledge, candidate-list artifacts, and target-row lexical signal can explain much of the apparent success.

\noindent The guarded synthetic tasks remove those shortcuts and give a sharper negative result.
On the first guarded synthetic task, the isolated control remains at chance, but the best cross-stream run is only 0.146 and the correct-answer probability remains essentially uniform.
Synthetic QA-v4 makes the same point more directly: a full-context SFT model reaches 0.984 accuracy when both evidence rows are placed in one sequence, while the isolated split model reaches 0.117 and the best split cross-stream variant reaches only 0.164.
We also tried reader variants that scored candidates stream-by-stream, message variants that placed short summaries in communication positions, and latent-memory variants with extra shared slots, but none closed the gap to full context.
By this standard, the current multi-hop QA results are short probes rather than final benchmarks.

\section{Additional MARL Architectures}
\label{app:marl-exploratory}

\noindent Table~\ref{tab:marl-exploratory} collects non-core MARL policy variants used as architecture probes but excluded from the main comparison set.

\begin{figure}[H]
\centering
\includegraphics[width=0.78\columnwidth]{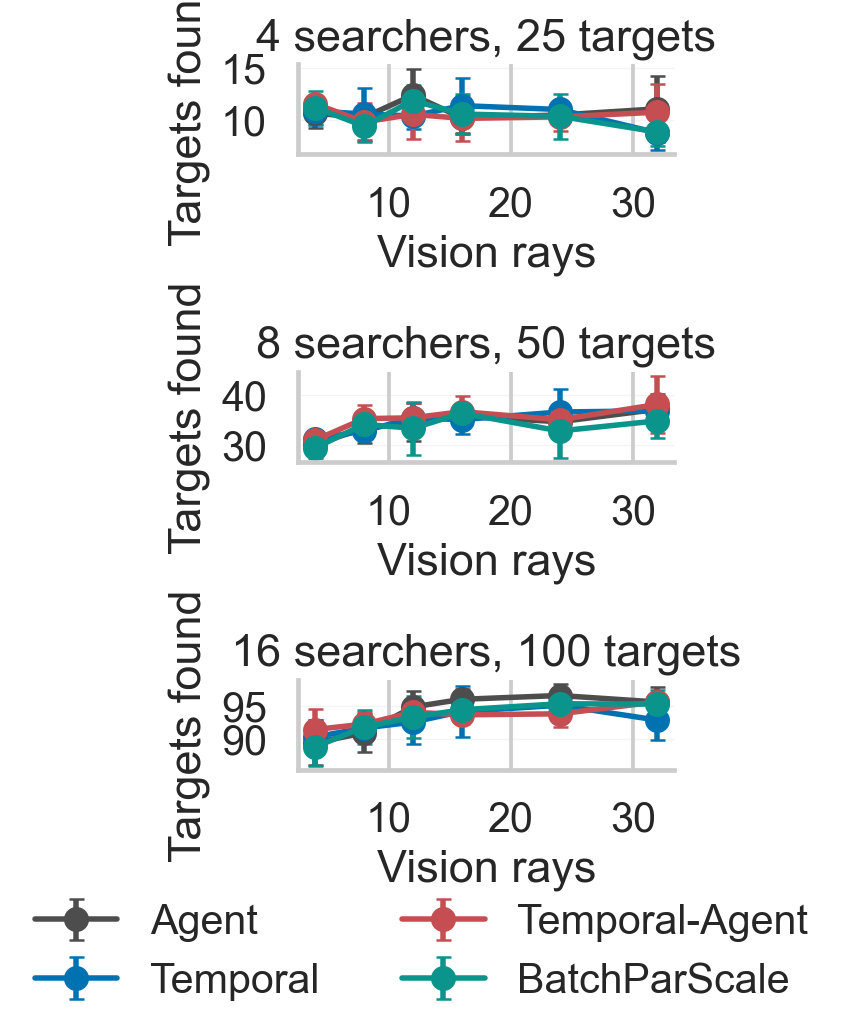}
\caption{SearchAndRescue env-size 4.0 visibility sweeps ordered by target count. Rows scale the number of searchers and targets together: 4 searchers with 25 targets, 8 searchers with 50 targets, and 16 searchers with 100 targets. Points report mean final targets found over five seeds, with error bars showing standard deviation. Higher is better.}
\label{fig:search-rescue-env4-sweeps}
\end{figure}

\begin{table*}[t]
\centering
\footnotesize
\setlength{\tabcolsep}{3pt}
\begin{tabular*}{\textwidth}{@{\extracolsep{\fill}}llrrrr@{}}
\toprule
Benchmark & Policy & Final & Std. / Win & Best & Last 10 / Time \\
\midrule
SMAX & Perceiver Transformer & 1.2 & 0.6 / 0.38 & \NA & 1580s \\
SMAX & Agent-Query Perceiver & 1.6 & 0.6 / 0.67 & \NA & 1418s \\
SMAX & Agent-Query Sink Perceiver & 1.2 & 0.6 / 0.39 & \NA & 1806s \\
SMAX & Agent-Query Sink Perceiver + RoPE & \textbf{2.0} & 0.0 / \textbf{0.96} & \NA & 1892s \\
SMAX & Sparse Cross-Batch Transformer & 0.7 & 0.0 / 0.02 & \NA & 1565s \\
MABrax & Last-token Agent Transformer & 90 & 13 & 106 & 94 \\
MABrax & Perceiver Transformer & 93 & 4 & 100 & 92 \\
\bottomrule
\end{tabular*}
\caption{Additional MARL architecture probes excluded from the main tables. SMAX entries use seeds 0, 1, and 2 from the A100 \texttt{num\_envs=64} rerun; their second metric reports return standard deviation / mean win rate, and the final column is mean wall-clock time per seed. MABrax entries report rollout-return standard deviation, best rollout return, and mean last-10 rollout return. \(\NA\) indicates that a baseline was not run for that suite. RoPE denotes rotary position embeddings \citep{su2024roformer}.}
\label{tab:marl-exploratory}
\end{table*}

\section{Default Cross-Stream Attention Mask}
\label{app:default-cross-stream-mask}

\noindent We define the mask used for the default sparse cross-stream attention pattern, omitting implementation switches for dropout, padding, and ablation modes.
Consider one original example expanded into \(B\) streams.
Within a layer \(\ell\), index a query by \((p,t)\) and a key-value by \((r,s)\), where \(p,r\in\{1,\ldots,B\}\) are stream indices and \(t,s\in\{1,\ldots,n\}\) are sequence positions.
Let \(C^{(\ell)}_p\subseteq\{1,\ldots,n\}\) be the communication positions selected for stream \(p\).
The Boolean allow-mask is defined by
\[
A^{(\ell)}_{(p,t),(r,s)} = 1
\]
if and only if either
\[
p=r \quad\text{and}\quad s\leq t,
\]
or
\[
p\neq r,\quad
t\in C^{(\ell)}_p,\quad
s\in C^{(\ell)}_r,\quad
s\leq t.
\]
Equivalently, same-stream tokens use standard causal self-attention, while cross-stream attention is available only from a communication query position to a communication key-value position in another stream, with the same causal order constraint.
The additive attention mask used in the softmax is
\[
M^{(\ell)}_{(p,t),(r,s)}
=
\begin{cases}
0, & A^{(\ell)}_{(p,t),(r,s)} = 1,\\
-\infty, & A^{(\ell)}_{(p,t),(r,s)} = 0.
\end{cases}
\]
For a static virtual-token model, \(C^{(\ell)}_p\) is the fixed set of virtual-token positions.
For a routed model, \(C^{(\ell)}_p\) is the layerwise top-\(K\) set selected by the router.

\subsection{FlexAttention Mask Implementation}
\label{app:flex-mask}

\noindent In FlexAttention, the mask is implemented as a Boolean \texttt{mask\_mod} rather than as a materialized additive mask \citep{attentiongym2024,dong2024flexattentionprogrammingmodel}.
We flatten the stream and sequence axes inside each communication group:
\[
\mathrm{flat}(p,t)=p\cdot n+t.
\]
Thus FlexAttention calls \texttt{mask\_mod} with flattened query and key-value indices.
The default sparse rule is implemented by recovering the stream and position indices, then returning the local causal rule or the cross-stream communication rule:

\begingroup
\footnotesize
\begin{verbatim}
def mask_mod(b, h, q_idx, kv_idx):
    q_stream = q_idx // seq_len
    q_pos = q_idx % seq_len
    kv_stream = kv_idx // seq_len
    kv_pos = kv_idx % seq_len

    same_stream = (
        q_stream == kv_stream
    )
    causal = kv_pos <= q_pos
    local = same_stream & causal

    q_selected = selected[
        b, q_stream, q_pos
    ]
    kv_selected = selected[
        b, kv_stream, kv_pos
    ]
    cross = (
        (~same_stream)
        & q_selected
        & kv_selected
        & causal
    )

    return local | cross
\end{verbatim}
\endgroup

\noindent Here \texttt{selected} has shape \((G,P,n)\), where \(G\) is the number of original examples in the batch after grouping, \(P\) is the number of streams, and \(n\) is the per-stream sequence length.
The tensors passed to FlexAttention have shape \((G,H,Pn,d_h)\), obtained by reshaping from \((G,P,H,n,d_h)\).
The full call is:

\begingroup
\footnotesize
\begin{verbatim}
q_flat = q.permute(
    0, 2, 1, 3, 4
).reshape(G, H, P * n, d_h)
k_flat = k.permute(
    0, 2, 1, 3, 4
).reshape(G, H, P * n, d_h)
v_flat = v.permute(
    0, 2, 1, 3, 4
).reshape(G, H, P * n, d_h)

block_mask = create_block_mask(
    mask_mod,
    B=G,
    H=H,
    Q_LEN=P * n,
    KV_LEN=P * n,
    device=q.device,
)

out_flat = flex_attention(
    q_flat,
    k_flat,
    v_flat,
    block_mask=block_mask,
)

out = out_flat.reshape(
    G, H, P, n, d_h
).permute(0, 2, 1, 3, 4)
\end{verbatim}
\endgroup

\noindent For static virtual tokens, \texttt{selected} is a fixed Boolean position mask broadcast across groups and streams.
For dynamic routing, \texttt{selected} is produced by a per-stream top-\(K\) router over sequence positions.

\bibliography{acl}

\end{document}

%% file: math_commands.tex
\newcommand{\method}{BatchParScale}
\newcommand{\xattn}{batch-mixing}
\newcommand{\Xattn}{Batch-mixing}
\newcommand{\parscale}{ParScale}

\newcommand{\pniah}{parallel-needle-in-a-haystack}

%% file: main.bbl
\begin{thebibliography}{43}
\providecommand{\natexlab}[1]{#1}

\bibitem[{Bahdanau et~al.(2015)Bahdanau, Cho, and Bengio}]{bahdanau2015neural}
Dzmitry Bahdanau, Kyunghyun Cho, and Yoshua Bengio. 2015.
\newblock Neural machine translation by jointly learning to align and
  translate.
\newblock In \emph{International Conference on Learning Representations}.

\bibitem[{Beltagy et~al.(2020)Beltagy, Peters, and
  Cohan}]{beltagy2020longformer}
Iz~Beltagy, Matthew~E. Peters, and Arman Cohan. 2020.
\newblock Longformer: The long-document transformer.
\newblock \emph{arXiv preprint arXiv:2004.05150}.

\bibitem[{Bonnet et~al.(2024)Bonnet, Luo, Byrne, Surana, Abramowitz, Duckworth,
  Coyette, Midgley, Tegegn, Kalloniatis, Mahjoub, Macfarlane, Smit, Grinsztajn,
  Boige, Waters, Mimouni, Mbou~Sob, de~Kock, Singh, Furelos-Blanco, Le,
  Pretorius, and Laterre}]{bonnet2024jumanji}
Cl{\'e}ment Bonnet, Daniel Luo, Donal Byrne, Shikha Surana, Sasha Abramowitz,
  Paul Duckworth, Vincent Coyette, Laurence~I. Midgley, Elshadai Tegegn,
  Tristan Kalloniatis, Omayma Mahjoub, Matthew Macfarlane, Andries~P. Smit,
  Nathan Grinsztajn, Raphael Boige, Cemlyn~N. Waters, Mohamed~A. Mimouni,
  Ulrich~A. Mbou~Sob, Ruan de~Kock, and 5 others. 2024.
\newblock Jumanji: A diverse suite of scalable reinforcement learning
  environments in {JAX}.
\newblock In \emph{International Conference on Learning Representations}.

\bibitem[{Brown et~al.(2024)Brown, Juravsky, Ehrlich, Clark, Le, R{\'e}, and
  Mirhoseini}]{brown2024largelanguagemonkeys}
Bradley Brown, Jordan Juravsky, Ryan Ehrlich, Ronald Clark, Quoc~V. Le,
  Christopher R{\'e}, and Azalia Mirhoseini. 2024.
\newblock \href {https://arxiv.org/abs/2407.21787} {Large language monkeys:
  Scaling inference compute with repeated sampling}.
\newblock \emph{Preprint}, arXiv:2407.21787.

\bibitem[{Busoniu et~al.(2008)Busoniu, Babu{\v{s}}ka, and
  De~Schutter}]{busoniu2008survey}
Lucian Busoniu, Robert Babu{\v{s}}ka, and Bart De~Schutter. 2008.
\newblock A comprehensive survey of multiagent reinforcement learning.
\newblock \emph{IEEE Transactions on Systems, Man, and Cybernetics, Part C
  (Applications and Reviews)}, 38(2):156--172.

\bibitem[{Chen et~al.(2025)Chen, Hui, Cui, Yang, Liu, Sun, Lin, and
  Liu}]{chen2025parscale}
Mouxiang Chen, Binyuan Hui, Zeyu Cui, Jiaxi Yang, Dayiheng Liu, Jianling Sun,
  Junyang Lin, and Zhongxin Liu. 2025.
\newblock \href {https://arxiv.org/abs/2505.10475} {Parallel scaling law for
  language models}.
\newblock \emph{arXiv preprint arXiv:2505.10475}.

\bibitem[{Child et~al.(2019)Child, Gray, Radford, and
  Sutskever}]{child2019sparse}
Rewon Child, Scott Gray, Alec Radford, and Ilya Sutskever. 2019.
\newblock Generating long sequences with sparse transformers.
\newblock \emph{arXiv preprint arXiv:1904.10509}.

\bibitem[{Cobbe et~al.(2021)Cobbe, Kosaraju, Bavarian, Chen, Jun, Kaiser,
  Plappert, Tworek, Hilton, Nakano, Ray, and
  Schulman}]{cobbe2021trainingverifiers}
Karl Cobbe, Vineet Kosaraju, Mohammad Bavarian, Mark Chen, Heewoo Jun, Lukasz
  Kaiser, Matthias Plappert, Jerry Tworek, Jacob Hilton, Reiichiro Nakano, Alex
  Ray, and John Schulman. 2021.
\newblock \href {https://arxiv.org/abs/2110.14168} {Training verifiers to solve
  math word problems}.
\newblock \emph{Preprint}, arXiv:2110.14168.

\bibitem[{Das et~al.(2019)Das, Gervet, Romoff, Batra, Parikh, Rabbat, and
  Pineau}]{das2019tarmac}
Abhishek Das, Theophile Gervet, Joshua Romoff, Dhruv Batra, Devi Parikh,
  Michael Rabbat, and Joelle Pineau. 2019.
\newblock {TarMAC}: Targeted multi-agent communication.
\newblock In \emph{International Conference on Machine Learning}.

\bibitem[{Dong et~al.(2024)Dong, Feng, Guessous, Liang, and
  He}]{dong2024flexattentionprogrammingmodel}
Juechu Dong, Boyuan Feng, Driss Guessous, Yanbo Liang, and Horace He. 2024.
\newblock \href {https://arxiv.org/abs/2412.05496} {Flex attention: A
  programming model for generating optimized attention kernels}.
\newblock \emph{Preprint}, arXiv:2412.05496.

\bibitem[{Eldan and Li(2023)}]{eldan2023tinystoriessmalllanguagemodels}
Ronen Eldan and Yuanzhi Li. 2023.
\newblock \href {https://arxiv.org/abs/2305.07759} {Tinystories: How small can
  language models be and still speak coherent english?}
\newblock \emph{Preprint}, arXiv:2305.07759.

\bibitem[{Freeman et~al.(2021)Freeman, Frey, Raichuk, Girgin, Mordatch, and
  Bachem}]{freeman2021brax}
C.~Daniel Freeman, Erik Frey, Anton Raichuk, Sertan Girgin, Igor Mordatch, and
  Olivier Bachem. 2021.
\newblock Brax: A differentiable physics engine for large scale rigid body
  simulation.

\bibitem[{Hsieh et~al.(2024)Hsieh, Sun, Kriman, Acharya, Rekesh, Jia, Zhang,
  and Ginsburg}]{hsieh2024ruler}
Cheng-Ping Hsieh, Simeng Sun, Samuel Kriman, Shantanu Acharya, Dima Rekesh, Fei
  Jia, Yang Zhang, and Boris Ginsburg. 2024.
\newblock \href {https://arxiv.org/abs/2404.06654} {{RULER}: What's the real
  context size of your long-context language models?}
\newblock In \emph{Conference on Language Modeling}.

\bibitem[{Jaegle et~al.(2022)Jaegle, Borgeaud, Alayrac, Doersch, Ionescu, Ding,
  Koppula, Zoran, Brock, Shelhamer, H{\'e}naff, Botvinick, Zisserman, Vinyals,
  and Carreira}]{jaegle2021perceiverio}
Andrew Jaegle, Sebastian Borgeaud, Jean-Baptiste Alayrac, Carl Doersch, Catalin
  Ionescu, David Ding, Skanda Koppula, Daniel Zoran, Andrew Brock, Evan
  Shelhamer, Olivier~J. H{\'e}naff, Matthew Botvinick, Andrew Zisserman, Oriol
  Vinyals, and Joao Carreira. 2022.
\newblock Perceiver {IO}: A general architecture for structured inputs and
  outputs.
\newblock In \emph{International Conference on Learning Representations}.

\bibitem[{Jaegle et~al.(2021)Jaegle, Gimeno, Brock, Zisserman, Vinyals, and
  Carreira}]{jaegle2021perceiver}
Andrew Jaegle, Felix Gimeno, Andy Brock, Andrew Zisserman, Oriol Vinyals, and
  Joao Carreira. 2021.
\newblock Perceiver: General perception with iterative attention.
\newblock In \emph{International Conference on Machine Learning}.

\bibitem[{Kamradt(2023)}]{kamradt2023needle}
Greg Kamradt. 2023.
\newblock Needle in a haystack: Pressure testing {LLMs}.
\newblock \url{https://github.com/gkamradt/LLMTest_NeedleInAHaystack}.

\bibitem[{Khot et~al.(2020)Khot, Clark, Guerquin, Jansen, and
  Sabharwal}]{khot2020qasc}
Tushar Khot, Peter Clark, Michal Guerquin, Peter Jansen, and Ashish Sabharwal.
  2020.
\newblock {QASC}: A dataset for question answering via sentence composition.
\newblock In \emph{AAAI Conference on Artificial Intelligence}, volume~34,
  pages 8082--8090.

\bibitem[{Kim et~al.(2026)Kim, Lee, Jang, Lee, and
  Lee}]{kim2026multisequenceverifiers}
Yegon Kim, Seungyoo Lee, Chaeyun Jang, Hyungi Lee, and Juho Lee. 2026.
\newblock \href {https://arxiv.org/abs/2603.03417} {Parallel test-time scaling
  with multi-sequence verifiers}.
\newblock \emph{Preprint}, arXiv:2603.03417.

\bibitem[{Lee et~al.(2019)Lee, Lee, Kim, Kosiorek, Choi, and
  Teh}]{lee2019settransformer}
Juho Lee, Yoonho Lee, Jungtaek Kim, Adam Kosiorek, Seungjin Choi, and Yee~Whye
  Teh. 2019.
\newblock Set transformer: A framework for attention-based
  permutation-invariant neural networks.
\newblock In \emph{International Conference on Machine Learning}.

\bibitem[{Lightman et~al.(2023)Lightman, Kosaraju, Burda, Edwards, Baker, Lee,
  Leike, Schulman, Sutskever, and Cobbe}]{lightman2023letsverifystep}
Hunter Lightman, Vineet Kosaraju, Yura Burda, Harri Edwards, Bowen Baker, Teddy
  Lee, Jan Leike, John Schulman, Ilya Sutskever, and Karl Cobbe. 2023.
\newblock \href {https://arxiv.org/abs/2305.20050} {Let's verify step by step}.
\newblock \emph{Preprint}, arXiv:2305.20050.

\bibitem[{Littman(1994)}]{littman1994markov}
Michael~L. Littman. 1994.
\newblock Markov games as a framework for multi-agent reinforcement learning.
\newblock In \emph{International Conference on Machine Learning}, pages
  157--163.

\bibitem[{Liu et~al.(2024)Liu, Lin, Hewitt, Paranjape, Bevilacqua, Petroni, and
  Liang}]{liu2024lostmiddle}
Nelson~F. Liu, Kevin Lin, John Hewitt, Ashwin Paranjape, Michele Bevilacqua,
  Fabio Petroni, and Percy Liang. 2024.
\newblock \href {https://doi.org/10.1162/tacl_a_00638} {Lost in the middle: How
  language models use long contexts}.
\newblock \emph{Transactions of the Association for Computational Linguistics},
  12:157--173.

\bibitem[{Lowe et~al.(2017)Lowe, Wu, Tamar, Harb, Abbeel, and
  Mordatch}]{lowe2017maddpg}
Ryan Lowe, Yi~Wu, Aviv Tamar, Jean Harb, Pieter Abbeel, and Igor Mordatch.
  2017.
\newblock Multi-agent actor-critic for mixed cooperative-competitive
  environments.
\newblock \emph{Advances in Neural Information Processing Systems}.

\bibitem[{Mahjoub et~al.(2025)Mahjoub, Abramowitz, de~Kock, Khlifi, du~Toit,
  Daniel, Nessir, Beyers, Clark, and Pretorius}]{mahjoub2025sable}
Omayma Mahjoub, Sasha Abramowitz, Ruan de~Kock, Wiem Khlifi, Simon du~Toit,
  Jemma Daniel, Louay~Ben Nessir, Louise Beyers, Liam Clark, and Arnu
  Pretorius. 2025.
\newblock Sable: a performant, efficient and scalable sequence model for marl.
\newblock In \emph{International Conference on Machine Learning}.

\bibitem[{Oliehoek et~al.(2008)Oliehoek, Spaan, and
  Vlassis}]{oliehoek2008decpomdp}
Frans~A. Oliehoek, Matthijs T.~J. Spaan, and Nikos Vlassis. 2008.
\newblock Optimal and approximate q-value functions for decentralized {POMDPs}.
\newblock \emph{Journal of Artificial Intelligence Research}, 32:289--353.

\bibitem[{Penedo et~al.(2024)Penedo, Kydl{\'i}{\v{c}}ek, Ben~Allal, Lozhkov,
  Mitchell, Raffel, von Werra, and Wolf}]{penedo2024fineweb}
Guilherme Penedo, Hynek Kydl{\'i}{\v{c}}ek, Loubna Ben~Allal, Anton Lozhkov,
  Margaret Mitchell, Colin Raffel, Leandro von Werra, and Thomas Wolf. 2024.
\newblock The {FineWeb} datasets: Decanting the web for the finest text data at
  scale.
\newblock In \emph{Advances in Neural Information Processing Systems}.

\bibitem[{{PyTorch Labs}(2024)}]{attentiongym2024}
{PyTorch Labs}. 2024.
\newblock {Attention Gym}: Helpful tools and examples for working with
  {FlexAttention}.
\newblock \url{https://github.com/pytorch-labs/attention-gym}.
\newblock GitHub repository.

\bibitem[{Raposo et~al.(2024)Raposo, Ritter, Richards, Lillicrap, Humphreys,
  and Santoro}]{raposo2024mixtureofdepthsdynamicallyallocatingcompute}
David Raposo, Sam Ritter, Blake Richards, Timothy Lillicrap, Peter~Conway
  Humphreys, and Adam Santoro. 2024.
\newblock \href {https://arxiv.org/abs/2404.02258} {Mixture-of-depths:
  Dynamically allocating compute in transformer-based language models}.
\newblock \emph{Preprint}, arXiv:2404.02258.

\bibitem[{Rutherford et~al.(2023)Rutherford, Ellis, Gallici, Cook, Lupu,
  Ingvarsson, Willi, Khan, de~Witt, and Foerster}]{rutherford2023jaxmarl}
Alexander Rutherford, Benjamin Ellis, Matteo Gallici, Justin Cook, Andrei Lupu,
  Gardar Ingvarsson, Timon Willi, Akbir Khan, Christian~Schroeder de~Witt, and
  Jakob Foerster. 2023.
\newblock \href {https://arxiv.org/abs/2311.10090} {Jaxmarl: Multi-agent rl
  environments and algorithms in jax}.
\newblock \emph{Preprint}, arXiv:2311.10090.

\bibitem[{Samvelyan et~al.(2019)Samvelyan, Rashid, de~Witt, Farquhar, Nardelli,
  Rudner, Hung, Torr, Foerster, and Whiteson}]{samvelyan2019smac}
Mikayel Samvelyan, Tabish Rashid, Christian~Schroeder de~Witt, Gregory
  Farquhar, Nantas Nardelli, Tim G.~J. Rudner, Chia-Man Hung, Philip H.~S.
  Torr, Jakob Foerster, and Shimon Whiteson. 2019.
\newblock The starcraft multi-agent challenge.
\newblock In \emph{International Conference on Autonomous Agents and Multiagent
  Systems}, pages 2186--2188.

\bibitem[{Schulman et~al.(2017)Schulman, Wolski, Dhariwal, Radford, and
  Klimov}]{schulman2017ppo}
John Schulman, Filip Wolski, Prafulla Dhariwal, Alec Radford, and Oleg Klimov.
  2017.
\newblock \href {https://arxiv.org/abs/1707.06347} {Proximal policy
  optimization algorithms}.
\newblock \emph{Preprint}, arXiv:1707.06347.

\bibitem[{Singh et~al.(2019)Singh, Jain, and Sukhbaatar}]{singh2019ic3net}
Amanpreet Singh, Tushar Jain, and Sainbayar Sukhbaatar. 2019.
\newblock Learning when to communicate at scale in multiagent cooperative and
  competitive tasks.
\newblock In \emph{International Conference on Learning Representations}.

\bibitem[{Snell et~al.(2024)Snell, Lee, Xu, and Kumar}]{snell2024testtime}
Charlie Snell, Jaehoon Lee, Kelvin Xu, and Aviral Kumar. 2024.
\newblock \href {https://arxiv.org/abs/2408.03314} {Scaling llm test-time
  compute optimally can be more effective than scaling model parameters}.
\newblock \emph{Preprint}, arXiv:2408.03314.

\bibitem[{Su et~al.(2024)Su, Lu, Pan, Murtadha, Wen, and Liu}]{su2024roformer}
Jianlin Su, Yu~Lu, Shengfeng Pan, Ahmed Murtadha, Bo~Wen, and Yunfeng Liu.
  2024.
\newblock {RoFormer}: Enhanced transformer with rotary position embedding.
\newblock \emph{Neurocomputing}, 568:127063.

\bibitem[{Sukhbaatar and Fergus(2016)}]{sukhbaatar2016commnet}
Sainbayar Sukhbaatar and Rob Fergus. 2016.
\newblock Learning multiagent communication with backpropagation.
\newblock In \emph{Advances in Neural Information Processing Systems}.

\bibitem[{Tan(1993)}]{tan1993multiagent}
Ming Tan. 1993.
\newblock Multi-agent reinforcement learning: Independent vs. cooperative
  agents.
\newblock In \emph{Proceedings of the Tenth International Conference on Machine
  Learning}, pages 330--337. Morgan Kaufmann.

\bibitem[{Vaswani et~al.(2017)Vaswani, Shazeer, Parmar, Uszkoreit, Jones,
  Gomez, Kaiser, and Polosukhin}]{vaswani2017attention}
Ashish Vaswani, Noam Shazeer, Niki Parmar, Jakob Uszkoreit, Llion Jones,
  Aidan~N. Gomez, Lukasz Kaiser, and Illia Polosukhin. 2017.
\newblock Attention is all you need.
\newblock \emph{Advances in Neural Information Processing Systems}.

\bibitem[{Wang et~al.(2023)Wang, Wei, Schuurmans, Le, Chi, Narang, Chowdhery,
  and Zhou}]{wang2023selfconsistency}
Xuezhi Wang, Jason Wei, Dale Schuurmans, Quoc Le, Ed~Chi, Sharan Narang,
  Aakanksha Chowdhery, and Denny Zhou. 2023.
\newblock Self-consistency improves chain of thought reasoning in language
  models.
\newblock In \emph{International Conference on Learning Representations}.

\bibitem[{Welbl et~al.(2018)Welbl, Stenetorp, and
  Riedel}]{welbl2018constructing}
Johannes Welbl, Pontus Stenetorp, and Sebastian Riedel. 2018.
\newblock Constructing datasets for multi-hop reading comprehension across
  documents.
\newblock In \emph{Transactions of the Association for Computational
  Linguistics}, volume~6, pages 287--302.

\bibitem[{Wen et~al.(2022)Wen, Kuba, Lin, Zhang, Wen, Wang, and
  Yang}]{wen2022mat}
Muning Wen, Jakub~Grudzien Kuba, Runji Lin, Weinan Zhang, Ying Wen, Jun Wang,
  and Yaodong Yang. 2022.
\newblock Multi-agent reinforcement learning is a sequence modeling problem.
\newblock In \emph{Advances in Neural Information Processing Systems}.

\bibitem[{Yang et~al.(2025)Yang, Li, Yang, Zhang, Hui, Zheng, Yu, Gao, Huang,
  Lv et~al.}]{yang2025qwen3}
An~Yang, Anfeng Li, Baosong Yang, Beichen Zhang, Binyuan Hui, Bo~Zheng, Bowen
  Yu, Chang Gao, Chengen Huang, Chenxu Lv, and 1 others. 2025.
\newblock \href {https://arxiv.org/abs/2505.09388} {Qwen3 technical report}.
\newblock \emph{Preprint}, arXiv:2505.09388.

\bibitem[{Yu et~al.(2022)Yu, Velu, Vinitsky, Wang, Bayen, and Wu}]{yu2022mappo}
Chao Yu, Akash Velu, Eugene Vinitsky, Yu~Wang, Alexandre Bayen, and Yi~Wu.
  2022.
\newblock The surprising effectiveness of ppo in cooperative multi-agent games.
\newblock In \emph{Advances in Neural Information Processing Systems}.

\bibitem[{Zaheer et~al.(2020)Zaheer, Guruganesh, Dubey, Ainslie, Alberti,
  Ontanon, Pham, Ravula, Wang, Yang, and Ahmed}]{zaheer2020bigbird}
Manzil Zaheer, Guru Guruganesh, Kumar~Avinava Dubey, Joshua Ainslie, Chris
  Alberti, Santiago Ontanon, Philip Pham, Anirudh Ravula, Qifan Wang, Li~Yang,
  and Amr Ahmed. 2020.
\newblock Big bird: Transformers for longer sequences.
\newblock In \emph{Advances in Neural Information Processing Systems}.

\end{thebibliography}
